\documentclass[sn-mathphys]{sn-jnl}

\usepackage{graphicx}%
\usepackage{multirow}%
\usepackage{amsmath,amssymb,amsfonts}%
\usepackage{amsthm}%
\usepackage{mathrsfs}%
\usepackage[title]{appendix}%
\usepackage{xcolor}%
\usepackage{textcomp}%
\usepackage{manyfoot}%
\usepackage{booktabs}%
\usepackage{algorithm}%
\usepackage{algorithmicx}%
\usepackage{algpseudocode}%
\usepackage{listings}%
\usepackage{pdflscape}
\usepackage{rotating}
\usepackage{graphicx}
\usepackage{tikz}
\usetikzlibrary{arrows.meta,positioning,shapes.geometric}
\usepackage{placeins}
\usepackage{longtable}
\usepackage{booktabs}  

\theoremstyle{thmstyleone}%
\theoremstyle{thmstyletwo}%

\theoremstyle{thmstylethree}%

\begin{document}

\title[Article Title]{Improved adaptive wind driven optimization algorithm for real-time path planning}


\author[1]{\fnm{Shiqian} \sur{Liu}}\email{liushiqian@gradate.utm.my}

\author[1]{\fnm{Azlan} \sur{Mohd Zain }}\email{azlanmz@utm.my }

\author[2]{\fnm{Le-le} \sur{Mao}}\email{maolele@hsnc.edu.cn }

\affil[1]{\orgdiv{Faculty of Computing}, \orgname{Universiti Teknologi Malaysia}, \orgaddress{\city{Skudai}, \postcode{81310}, \state{Johor}, \country{Malaysia}}}

\affil[2]{\orgdiv{College of Mathematics and Computer Science}, \orgname{Hengshui University}, \orgaddress{\street{Street}, \city{Hengshui}, \postcode{053000}, \state{Hebei}, \country{China}}}


\abstract{Recently, path planning has achieved remarkable progress in enhancing global search capability and convergence accuracy through heuristic and learning-inspired optimization frameworks. However, real-time adaptability in dynamic environments remains a critical challenge for autonomous navigation, particularly when robots must generate collision-free, smooth, and efficient trajectories under complex constraints. By analyzing the difficulties in dynamic path planning, the Wind Driven Optimization (WDO) algorithm emerges as a promising framework owing to its physically interpretable search dynamics. Motivated by these observations, this work revisits the WDO principle and proposes a variant formulation, Multi-hierarchical adaptive wind driven optimization(MAWDO), that improves adaptability and robustness in time-varying environments. To mitigate instability and premature convergence, a hierarchical-guidance mechanism divides the population into multiple groups guided by individual, regional, and global leaders to balance exploration and exploitation. Extensive evaluations on sixteen benchmark functions show that MAWDO achieves superior optimization accuracy, convergence stability, and adaptability over state-of-the art metaheuristics. In dynamic path planning, MAWDO shortens the path length to 469.28 pixels, improving over Multi-strategy ensemble wind driven optimization(MEWDO), Adaptive wind driven optimization(AWDO) and WDO by 3.51\%, 11.63\% and 14.93\%, and achieves the smallest optimality gap (1.01) with smoothness 0.71 versus 13.50 and 15.67 for AWDO and WDO, leading to smoother, shorter, and collision-free trajectories that confirm its effectiveness for real-time path planning in complex environments.}




\maketitle

\section{Introduction}\label{sec1}

Mobile robots are versatile platforms that integrate environment perception, dynamic decision-making, and behavior control. They are widely deployed in space exploration, emergency response, smart living, and defense applications \cite{siegwart2011, siciliano2016handbook}. Path planning, an NP-hard problem, is a core capability that reflects autonomous intelligence \cite{lavalle2006planning}. It computes a feasible, collision-free trajectory from a start to a target under kinematic and environmental constraints. By obstacle behavior, planning is categorized as static (fixed obstacles) or dynamic (time-varying obstacles) \cite{jaillet2010dynamic}; by map knowledge, it is global (leveraging prior maps) or local (constructed online from onboard sensing) \cite{choset2005}.

Safe, efficient navigation demands planners that are both fast and reliable because trajectory quality directly affects mission success and energy consumption. Existing approaches span classical search and sampling to modern learning-based methods. Representative techniques include Dijkstra’s algorithm \cite{dijkstra1959}, A* search \cite{a_star1968}, visibility graphs \cite{choset2005}, rapidly-exploring random trees (RRT) \cite{rrt2000}, artificial potential fields (APF) \cite{khatib1986apf}, and bug-style algorithms \cite{bug1987}. However, the inherently sequential, node-by-node expansion of graph search (Dijkstra and A*) causes steep growth in runtime and memory as map scale or clutter increases, undermining real-time performance and largely confining these methods to static scenes \cite{lavalle2006planning}. RRT offers only probabilistic completeness and often yields tortuous paths; APF is susceptible to local minima; and bug strategies can incur long detours as environmental complexity rises \cite{choset2005}. Learning-based planners, including deep learning (DL) and reinforcement learning (RL), offer adaptability under uncertainty \cite{levine2016jmlr, zhu2020iclr, radosavovic2022}, yet they remain challenged by training instability, infeasible motion predictions with incomplete data, and state-space explosion under limited compute \cite{tai2017rlnav}. These issues collectively impede strict real-time deployment on embedded platforms.

To overcome these limitations, meta-heuristic (swarm-intelligence) algorithms have been extensively adopted due to their population-based cooperative search, which enables efficient approximation of near-optimal solutions in complex and nonlinear landscapes \cite{engelbrecht2007, yang2021book}. Representative techniques such as genetic algorithms (GA) \cite{goldberg1989}, particle swarm optimization (PSO) \cite{kennedy1995pso}, ant colony optimization (ACO) \cite{dorigo2004aco}, whale optimization algorithm (WOA) \cite{mirjalili2016woa}, Eurasian Lynx Optimizer (ELO) \cite{dhiman2018elo}, and salp swarm algorithm (SSA) \cite{mirjalili2017ssa} have been successfully applied to both static and dynamic planning problems \cite{mei2021robotica}. In static environments, the main objective typically emphasizes shortest-path efficiency, whereas in dynamic scenarios, feasibility and safety constraints, such as collision avoidance during motion, take precedence, often at the expense of path length or smoothness \cite{jaillet2010dynamic}. As obstacle density and spatial complexity increase, algorithmic performance factors such as convergence reliability, diversity preservation, and computational latency become decisive \cite{engelbrecht2007}. Therefore, there remains strong demand for algorithms with enhanced environmental adaptability and real-time responsiveness.

WDO algorithm is particularly attractive for its low parameterization and implementation simplicity \cite{bayraktar2010wdo}. It demonstrates strong global exploration capability but remains vulnerable to premature convergence and accuracy degradation in multi-modal landscapes. To alleviate these shortcomings, numerous WDO variants have been proposed, including AWDO \cite{das2013awdo}, quantum-encoded QWDO \cite{qwdopaper}, Chan-QWDO \cite{chanqwdo2020}, binary adaptive WDO (iBAWDO) \cite{binaryawdo2021}, chaos–tanh and T-distribution mutation-based CHTWDO \cite{chtwdo2022}, and MEWDO \cite{mewdo2025robotpath}. In robotics, WDO and its variants have been applied to waypoint optimization, steering-angle control, and specialized trajectory generation \cite{mei2021robotica}. However, most existing studies remain confined to single-robot settings with limited obstacle complexity or focus on offline/global optimization. Achieving robust, real-time path planning in dynamic and computationally constrained environments continues to be an open and pressing challenge.

Path planning in dynamic, cluttered maps is inherently a multi-local-extremum optimization problem in which naive exploitation causes premature convergence and brittle behavior. In particular, standard WDO/AWDO formulations may over-amplify global guidance, collapse diversity, and become sensitive to low-dimensional path encodings, resulting in obstacle entrapment, long detours, and unstable convergence. Addressing these pain points requires (1) diversity-preserving exploration across distinct regions, (2) iteration-aware guidance schedules that gradually favor exploitation, and (3) low-latency inner loops and robust boundary handling to sustain real-time updates.

To this end, we propose the MAWDO algorithm. The primary contributions of this work are summarized as follows: (1) A multi-hierarchical AWDO framework is introduced to preserve cross-region diversity and mitigate premature convergence in multi-modal landscapes \cite{das2013awdo, oblpaper}. (2) A hierarchical-guidance strategy is developed, integrating personal, group, and global leaders through iteration-adaptive weighting. (3) A scheduled-mixing scheme coupled with a distance-gated control mechanism is proposed to balance local and global learning behaviors, accelerating convergence while maintaining solution quality. (4) Periodic Guided Restart (PGR) \cite{pgr2020} and Opposition-Based Learning (OBL) \cite{oblpaper} are incorporated to strengthen global search capability. (5) A collision-aware objective function is formulated by combining path length with a higher-order obstacle-intersection penalty, promoting both safety and smoothness. Extensive evaluations on dynamic environments show that MAWDO generates shorter, smoother, and more feasible paths with lower planning latency compared with representative benchmark algorithms. 

The remainder of this paper is organized as follows. Section 2 reviews the fundamentals of the WDO/AWDO family and establishes the path-encoding and cost model used in our planner. Section 3 details the proposed MAWDO framework. Section 4 presents the real-time dynamic path planning setup, objective design, and comparative experiments against classical and swarm-based baselines. Finally, Section 5 concludes the paper and discusses future directions such as multi-robot extension and hardware-in-the-loop deployment.

\section{Basic of algorithm}\label{sec2}

\subsection{Wind driven optimization}\label{subsec2}
WDO algorithm is a population-based metaheuristic inspired by the dynamic motion of air parcels in the Earth’s atmosphere. It models the natural process of atmospheric circulation, in which air parcels move under the influence of pressure gradients, gravitational forces, and Coriolis effects. In this analogy, each candidate solution corresponds to an air parcel whose position represents a possible solution in the search space, and its velocity reflects the rate of positional change \cite{bayraktar2010wdo}.

The WDO algorithm begins by initializing a population of $N$ air parcels within the predefined bounds of the search domain. Each parcel $i$ in dimension $d$ has an initial position $X_{id}$ and velocity $U_{id}$. The fitness of each parcel is evaluated according to the objective function $f(x_i)$, and the population is ranked based on ascending fitness values. The fundamental velocity update equation in WDO is derived from simplified physical laws of motion \cite{bayraktar2010wdo} in the atmosphere at iteration $t$, expressed as:
\begin{equation}
U_{i,d}^{t+1}=(1-\alpha)U_{i,d}^t-gX_{i,d}^t+RT(B_d^t-X_{i,d}^t)+cU_{i,d_{perm}}^t
\end{equation}
where $\alpha$ denotes the friction coefficient controlling inertia, $g$ represents the gravitational constant directing particles toward the origin, $RT$ is the pressure gradient coefficient driving particles toward the global best position $B$, and $c$ is the Coriolis coefficient simulating rotational deflection. The final term $U_{i,d_{perm}}$ introduces cross-dimensional coupling by randomly permuting the velocity components to simulate rotational turbulence\cite{bayraktar2010wdo,das2013awdo}. After updating the velocity, the new position is computed as:
\begin{equation}\label{eq:position}
    X_{i,d}^{t+1}=X_{i,d}^t+U_{i,d}^{t+1}
\end{equation}
To ensure that each position component remains within the predefined search limits, a boundary control strategy is employed after every position update. 
Specifically, the updated position vector is constrained by the following clamping rule: \begin{equation}
    X_i^{t+1} = \min \big( \max ( X_i^{t+1}, L_b ), U_b \big)
\end{equation}
where $L_b$ and $U_b$ denote the lower and upper bounds of the search space, respectively. This operation guarantees that no particle exceeds the feasible domain, thereby maintaining the stability and validity of the search process\cite{engelbrecht2007,yang2021book}.

The primary advantage of the WDO lies in its physical interpretability, low parameter sensitivity, and simple implementation structure. The algorithm relies on four main control coefficients, the friction coefficient $\alpha$, gravitational constant $g$, pressure gradient coefficient $RT$, and the Coriolis coefficient $c$, which together balance the trade-off between exploration and exploitation \cite{bayraktar2010wdo}. Because of its minimal parameterization, WDO can be easily adapted to diverse optimization problems without extensive empirical tuning, making it computationally efficient compared with other population-based algorithms such as Particle Swarm Optimization (PSO)\cite{kennedy1995pso}, Differential Evolution (DE) \cite{goldberg1989}, and other nature-inspired optimizers \cite{yang2021book,mirjalili2016woa}.

WDO algorithm is a population-based metaheuristic inspired by the dynamic motion of air parcels in the Earth’s atmosphere. It models the natural process of atmospheric circulation, in which air parcels move under the influence of pressure gradients, gravitational forces, and Coriolis effects. In this analogy, each candidate solution corresponds to an air parcel whose position represents a possible solution in the search space, and its velocity reflects the rate of positional change \cite{bayraktar2010wdo}.

Variants of WDO have further enhanced its performance, including Adaptive WDO \cite{das2013awdo}, Quantum-Inspired WDO \cite{qwdopaper}, Chaotic Quantum WDO \cite{chanqwdo2020}, OBL-based hybrid variants \cite{oblpaper}, and multi-strategy enhanced WDO models \cite{mewdo2025robotpath}, which improve convergence behavior and robustness in high-dimensional search spaces.

\subsection{Adaptive wind driven optimization}\label{subsec2}

AWDO algorithm is an enhanced variant of the standard WDO, designed to improve global search reliability and convergence accuracy. While the canonical WDO employs fixed physical coefficients that remain constant throughout the search \cite{bayraktar2010wdo}, AWDO introduces adaptive and stochastic parameter dynamics that enable the algorithm to self-regulate its exploration and exploitation tendencies \cite{das2013awdo}. By allowing the main coefficients—the air resistance $\alpha$, gravitational constant $g$, rate of ascent $RT$, and Coriolis effect coefficient $c$—to vary adaptively, AWDO achieves a more flexible balance between global exploration and local refinement \cite{das2013awdo}.

The WDO algorithm is governed by deterministic equations derived from atmospheric motion principles \cite{bayraktar2010wdo}. However, its convergence characteristics are sensitive to manual tuning of parameters $(\alpha, g, RT, c)$, which may lead to either premature convergence or excessive wandering \cite{yang2021book}. AWDO alleviates this issue by introducing adaptive randomization of these coefficients in each iteration \cite{das2013awdo}, thereby dynamically modulating the swarm’s kinetic behavior. This strategy ensures that the population experiences periodic changes in exploration intensity and convergence pressure, enabling more effective traversal of complex, multimodal landscapes without relying on predefined parameter schedules \cite{engelbrecht2007}.

Let $X_i^t \in \mathbb{R}^D$ and $U_i^t \in \mathbb{R}^D$ denote the position and velocity of the $i^{th}$ air parcel at iteration $t$. The AWDO update mechanism is described by the following equation:
\begin{equation}\label{eq:velocity}
    U_i^{t+1} = (1 - \alpha_t)U_i^t - g_t X_i^t + \frac{RT_t(B^t - X_i^t)}{r_i} + \frac{c_t U_{i,{perm}}^t}{r_i}
\end{equation}
where $\alpha_t, g_t, RT_t, c_t$ are adaptive coefficients sampled from a uniform distribution within $[0, 1]$ \cite{das2013awdo}. Here, $r_i$ denotes the rank index based on fitness, $B^t$ is the global best solution, and $U_{i,perm}^t$ represents the permuted velocity vector, introducing lateral movement diversity via the Coriolis effect \cite{bayraktar2010wdo}. This adaptive randomization produces a dynamic search pressure that prevents premature stagnation in local minima \cite{chanqwdo2020, mewdo2025robotpath}.

AWDO retains the same atmospheric metaphor as WDO but introduces adaptive stochastic control over the main parameters. Unlike standard WDO, where search dynamics remain fixed, AWDO allows the swarm to experience diverse aerodynamic conditions, simulating variable turbulence and wind behavior \cite{das2013awdo}.

However, AWDO still relies on full velocity permutation in the Coriolis term, which can induce instability in high-dimensional spaces \cite{qwdopaper}. Additionally, its boundary control mechanism remains a simple clamping operation, which may cause stagnation near the limits of the search domain. These limitations have motivated subsequent extensions, such as hierarchical guidance and simplified Coriolis design \cite{chtwdo2022}, reflection-with-damping boundaries, and multi-strategy hybridization to improve convergence stability and diversity \cite{mewdo2025robotpath, binaryawdo2021}.

\section{Improvement strategy}\label{sec3}

\subsection{Hierarchical Guidance }\label{subsec2}
In the AWDO algorithm, particle velocity updates rely heavily on the global best solution. This reliance often results in unstable search behavior: once a particle discovers a seemingly optimal point, it quickly becomes the global leader, causing all particles to converge toward the same location. Such premature and overly aggressive aggregation reduces exploration ability and may misguide the population toward suboptimal regions, ultimately compromising convergence performance and robustness.

To address this limitation, we introduce a hierarchical guidance strategy that partitions the population into eight independent groups, enabling distributed exploration across multiple regions of the search space. A composite guidance vector $B$ is then constructed through a weighted integration of three hierarchical sources, individual, regional, and global guidance, achieving a more balanced trade-off between exploration and exploitation.
\begin{equation}\label{eq:guidance}
    B = w_1 pbest + w_2 gbest_g + w_3 tbest, \qquad w_1 + w_2 + w_3 = 1
\end{equation}
In each iteration of the algorithm, all particles first evaluate their fitness values based on their current positions. Each particle then updates its personal best $pbest$ if its new fitness is better than its historical record. Afterward, all particles within a group are sorted by fitness, and the best-performing particle is designated as the group leader, with its position stored as the group best $gbest$. Finally, all groups are compared, and the best $gbest$ among them is selected as the global best $tbest$ corresponding to the lowest fitness value, which represents the overall leading solution in the population at that iteration. $w_1$, $w_2$, $w_3$ control the weight of $pbest$, $gbest$, $tbest$.

$B$ is built using those three values each iteration to form the weighted direction target for every particle. It’s the heart of the hierarchical guidance, transforming AWDO from a single-level optimizer into a multi-hierarchical adaptive search system. Particle velocities $U$ are updated in Eq.~\eqref{eq:velocity}, integrating multiple forces such as inertia, gravity, pressure gradient, and Coriolis effects. Finally, each particle’s position is updated in Eq.~\eqref{eq:position}, moving it to a new location that serves as the input for the next iteration.

\subsection{Dual-layer control }\label{subsec2}
\subsubsection{Scheduled Mixing Strategy}\label{subsubsec2}
For the weight of B formula in Eq.~\eqref{eq:guidance}, if we take the fixed weight, The algorithm might start focusing on the global best $tbest$ too soon, before enough of the search space has been explored. This increases the risk of premature convergence, getting stuck in a local optimum. If the weights are tuned to favor exploration high $w_1$, then later in the run the algorithm may still wander randomly instead of fine-tuning near the best region, resulting in slow or unstable convergence
To adaptively balance exploration and exploitation, we propose the scheduled mixing strategy. 
\begin{align}
w_1 &= 0.6(1-\lambda) + 0.2\lambda \\
w_2 &= 0.3(1-\lambda) + 0.3\lambda \\
w_3 &= 0.1(1-\lambda) + 0.5\lambda \\
\lambda &= \frac{t-1}{T-1}
\end{align}
The scheduled mixing strategy adjusts the three guidance weights $w_1$, $w_2$, $w_3$ dynamically over time to create a smooth transition from exploration to exploitation. In early iterations, a larger $w_1$ emphasizes each particle’s personal experience $pbest$, promoting wide exploration of the search space. As iterations progress, $w_3$ gradually increases, giving greater influence to the global best $tbest$ and encouraging convergence toward optimal regions. Meanwhile, $w_2$ remains moderate to maintain coordination among group members through the group best $gbest$. This adaptive scheduling prevents particles from converging too early on suboptimal solutions, balances local and global learning, and ultimately enhances both convergence speed and solution quality by ensuring a gradual, stable shift from exploration to exploitation.

\subsubsection{Distance-Gated Control}\label{subsubsec2}
In the hierarchical guidance mechanism, each group maintains a regional best solution $gbest$, while the algorithm also tracks a global best $tbest$. If a uniform global guidance weight $w_3$ is applied to all groups, every group is subjected to the same attraction toward the current global leader, irrespective of their spatial distance from it. This uniform pull can be detrimental for two reasons:  (1). Premature population convergence. Groups located near the global best are excessively reinforced, causing the entire population to rapidly collapse into a single region and increasing the risk of stagnation at a local optimum. (2). Loss of exploration diversity. Groups situated farther from the global best, which may hold the potential to explore promising alternative regions, are prematurely dragged toward the current optimum, undermining the multi-group architecture by forcing a single-centered search behavior.

To mitigate this issue, we introduce a spatial gating strategy that adaptively modulates the strength of global influence based on each group’s distance from the global best.
The distance-gate multiplies that baseline $w_3$ by a factor:
\begin{equation}
    gate_g = \min\left(1, \frac{\lVert gbest - tbest \rVert}{d_{\max}}\right)
\end{equation}
and produces a group-specific effective weight:
\begin{equation}
    w_3^{(eff)} = w_3\times gate_g
\end{equation}
The term $d_{max}$ represents the maximum possible distance between $gbest$ and $tbest$ within the search space. 
When the distance between $gbest$ and $tbest$ approaches the maximum range $d_{max}$, the gating factor $gate_g$ tends toward unity $(gate_g \approx 1)$. In this case, the global guidance weight $w_3$ remains strong, effectively pulling that group toward the global leader and enhancing convergence consistency across subpopulations. Conversely, when the group’s leader is already near the global best, $gate_g$ becomes significantly smaller $(gate_g \ll 1)$, thereby suppressing $w_3$ and allowing the group to perform localized exploration rather than collapsing prematurely around the global solution.

Through this mechanism, the proposed distance-gate introduces a form of spatial adaptivity that complements the existing temporal scheduling, enabling a balanced transition between global cooperation and local refinement throughout the optimization process.

\subsection{Periodic Guided Restart and Opposition-Based Learning}\label{subsec2}

The AWDO algorithm focused primarily on adaptive parameter control and rank-based velocity adjustment but lacked mechanisms for population rejuvenation. Consequently, it was prone to stagnation in complex multimodal landscapes. In this work, two complementary strategies, PGR and OBL, are introduced to overcome this limitation. PGR periodically regenerates a subset of inferior particles around the global best to restore search diversity, while OBL evaluates mirrored candidates to ensure global coverage. These additions substantially enhance AWDO’s ability to escape local optima and maintain balanced exploration–exploitation dynamics.
\subsubsection{PGR}\label{subsubsec2}
To mitigate population stagnation and premature convergence, a PGR mechanism is introduced. At fixed intervals, a small proportion of the worst-performing particles are reinitialized in the neighborhood of the current global best $tbest$. 
\begin{equation}\label{eq:PGR}
    X_{\text{new}} \sim \mathcal{N}\!\left(tbest, \sigma^{2}\right)
\end{equation}

Specifically, these particles are regenerated using a Gaussian distribution centered at $tbest$ with a standard deviation that gradually decreases over iterations $\sigma$-decay.
\begin{equation}
    \sigma_t = \sigma_0 - \alpha \cdot \frac{t}{T}
\end{equation}
This strategy maintains an adaptive balance between exploration and exploitation: early restarts encourage broad search diversity, whereas later restarts concentrate sampling near promising regions. Consequently, PGR rejuvenates the swarm by reintroducing exploratory candidates without disrupting the convergence trajectory of the leading particles.

\subsubsection{OBL}\label{subsubsec2}
To complement the guided restart, OBL is incorporated to enhance global search capability. For each newly generated particle $X_i$, an opposite candidate is:
\begin{equation}\label{eq:OBL}
    X'=lb+ub-X
\end{equation}
which is simultaneously produced within the same search range. Both $X_i$ and $X'_i$ are evaluated, and the superior one is retained for subsequent evolution. This symmetric exploration allows the algorithm to probe the opposite region of the search space, effectively doubling the sampling coverage and increasing the probability of escaping local optima.

\subsection{Boundary Reflection and Damping}\label{subsec2}
In the AWDO algorithm, boundary violations were handled by simple truncation, which directly clipped the particle position to the nearest bound. Although this method ensures feasibility, it often leads to the loss of velocity information and causes premature stagnation near the search boundaries. To address this limitation, a reflective-damped boundary mechanism is introduced.
When a particle exceeds a boundary, its position and velocity are updated according to:
\begin{equation}\label{eq:boundary}
    X \Leftarrow 2b-X
\end{equation}
\begin{equation}
    U \Leftarrow -\eta U
\end{equation}
where $b$ denotes the violated boundary (either the upper or lower limit), and $\eta$ is a damping coefficient (typically 0.5). This formulation mirrors the particle back into the feasible region while reversing and attenuating its velocity.
By preserving the kinetic continuity of the velocity term while constraining the position within the valid domain, this approach maintains the physical consistency of WDO-based dynamics and prevents the population from clustering along the search limits. Consequently, the reflective-damped strategy enhances convergence stability and promotes smoother boundary behavior compared with the traditional hard-clipping approach.

\subsection{Low dimensions}\label{subsec2}
\subsubsection{Heterogeneous Coefficients}\label{subsubsec2}
In high-dimensional spaces naturally promote directional diversity through near-orthogonal movement, making coefficient homogeneity far less detrimental and often more stable for convergence. 
In contrast, in low-dimensional spaces, particle trajectories tend to align due to limited geometric freedom, causing populations with identical AWDO coefficients to rapidly synchronize and collapse into the same local basin. 

To address this limitation, we introduce a heterogeneous-coefficients strategy, where each group independently samples a unique set of coefficients:
\begin{align}\label{eq:coefficient}
    a &\sim U(0.15, 0.45), \\
    g &\sim U(0.15, 0.45), \\
    RT &\sim U(0.90, 1.50), \\
    c &\sim U(0.10, 0.50)
\end{align}
As a result, different groups exhibit distinct behavioral tendencies: groups with higher RT and lower a intensify global exploitation and accelerate toward promising basins, while groups with higher damping (a) maintain cautious, fine-grained local search. This controlled behavioral diversity breaks trajectory synchronization, enabling parallel exploration of multiple regions of the search space.

Overall, this mechanism significantly stabilizes the optimization dynamics in low-dimensional problems by preserving population diversity while maintaining sufficient exploitation pressure, thereby improving global optimality and reducing the probability of entrapment in local minima.

\subsubsection{Velocity Limit Tightening}\label{subsubsec2}
The AWDO formulation allows relatively large velocity magnitudes, which is suitable for exploration in high-dimensional landscapes but often leads to step-size overshoot and oscillatory behavior in low-dimensional spaces. To address this instability, the proposed approach introduces a velocity-limit tightening mechanism activated when (D <= 3). Specifically, the velocity upper bound is reduced to one-quarter of the nominal limit derived from the search range. 
\begin{equation}\label{eq:tighten}
    v_{max} \Leftarrow 0.25v_{max}
\end{equation}
This constraint maintains adequate search mobility while significantly improving numerical stability by preventing abrupt position jumps, excessive boundary interactions, and chaotic velocity reversals. Consequently, the particles follow smoother convergence trajectories and achieve more reliable exploitation performance.

\subsubsection{Centroid-Based Gravitational Stabilization}\label{subsubsec2}
In AWDO, the gravitational term attracts all particles toward the coordinate origin, which can introduce an unnecessary spatial bias and cause early population collapse. To overcome this, the proposed method replaces the static origin with the dynamic centroid of each group. Each particle is therefore attracted toward its group’s mean position, expressed as:
\begin{equation}
    grav=-g(X-\bar{X})
\end{equation}
\begin{equation}
    \bar{X}=\frac{1}{N} \sum_{n=1}^{N} X_n
\end{equation}
This centroid-based attraction enhances local stability and prevents uncontrolled divergence in low-dimensional search spaces while preserving the independence of different groups, thereby improving exploration balance and convergence reliability.

\subsection{Algorithmic processes}\label{subsec2}
\begin{figure}[htbp]
\centering
\resizebox{0.9\textwidth}{!}{
\begin{tikzpicture}[
    node distance=0.9cm,
    every node/.style={font=\small},
    startstop/.style = {ellipse, draw, minimum width=1.8cm, minimum height=0.8cm},
    process/.style    = {rectangle, draw, minimum width=3.8cm, minimum height=0.8cm,
                         text width=5cm, align=center},
    decision/.style   = {diamond, draw, aspect=2, inner sep=1pt,
                         text width=2.6cm, align=center},
    arrow/.style      = {->, >=Stealth}
]

\node[startstop] (start) {Start};

\node[process, below=of start] (init) {Input parameters $(N,D,lb,ub,T_{\max},t)$\\
Initialize positions $X$, velocities $U$\\
Partition population into 8 groups};

\node[decision, below=of init] (tmax) {$t < T_{\max}$?};

\node[decision, below=of tmax] (dim) {$D > 3$?};

\node[process, below=of dim] (best) {Calculate individual fitness value\\
Get \textit{pbest} of personal best\\
Get \textit{gbest} within each group\\
Get \textit{tbest} across groups};

\node[process, below=of best] (updateB) {Update guidance vector $B$ according to Eq.~\eqref{eq:guidance}\\
Update velocity $U$ according to Eq.~\eqref{eq:velocity}\\
Update position $X$ according to Eq.~\eqref{eq:position}};

\node[process, below=of updateB] (updateX) {Update the position $X$ according to Eq.~\eqref{eq:PGR}, Eq.~\eqref{eq:OBL}, Eq.~\eqref{eq:boundary}\\
};

\node[process, below=of updateX] (tplus) {$t = t + 1$};

\node[process, right=2.8cm of tmax] (output) {Output \textit{tbest} and best fitness};

\node[startstop, right=2cm of output] (end) {End};

\node[process, right=2.8cm of dim] (coeff) {Assign coefficients according to Eq.~\eqref{eq:coefficient}\\
Tighten $v_{\max}$ according to Eq.~\eqref{eq:tighten}};

\draw[arrow] (start)  -- (init);
\draw[arrow] (init)   -- (tmax);
\draw[arrow] (tmax)   -- node[right]{Yes} (dim);
\draw[arrow] (dim)    -- node[right]{Yes} (best);
\draw[arrow] (best)   -- (updateB);
\draw[arrow] (updateB)-- (updateX);
\draw[arrow] (updateX)-- (tplus);

\draw[arrow] (tplus.west) -| ([xshift=-1.2cm]tmax.west) |- (tmax.west);

\draw[arrow] (tmax.east) -- node[above]{No} (output.west);
\draw[arrow] (output) -- (end);

\draw[arrow] (dim.east) -- node[above]{No} (coeff.west);
\draw[arrow] (coeff.south) |- (best.east);

\end{tikzpicture}
}
\caption{Flowchart of the proposed algorithm}
\label{fig:proposed_algorithm}
\end{figure}
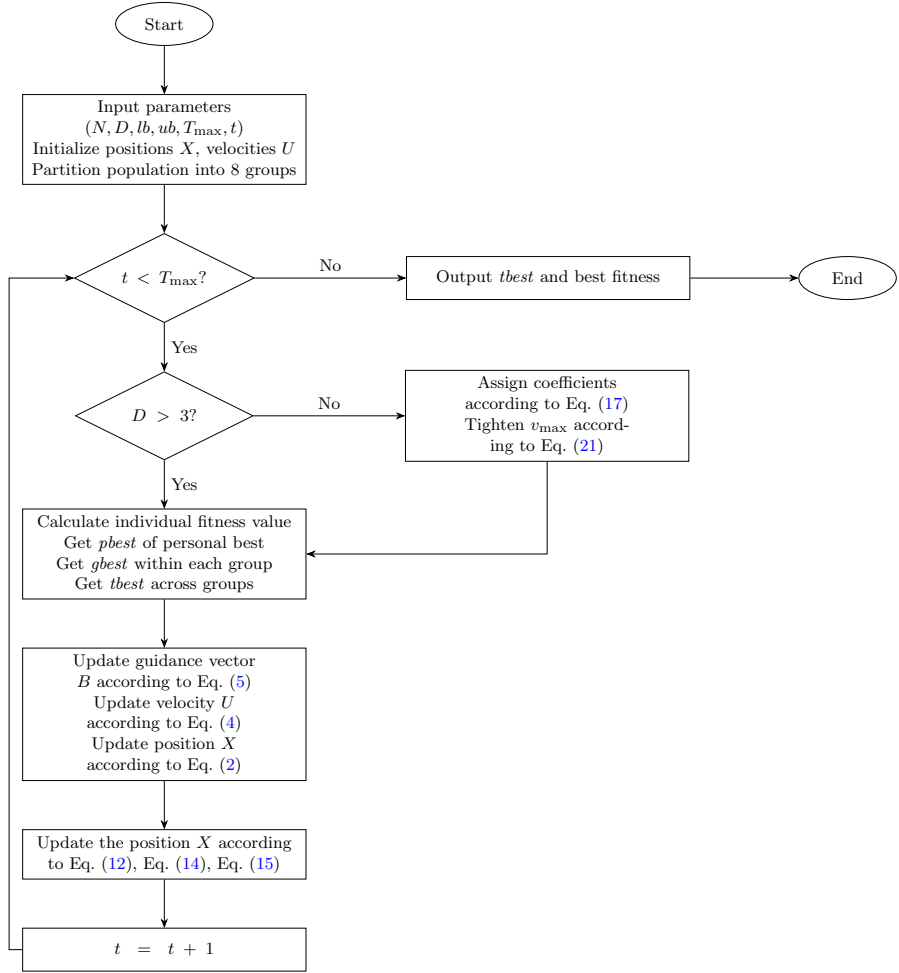

The proposed MAWDO framework enhances AWDO through a coherent set of improvement strategies targeting exploration-exploitation balance, stagnation avoidance, and numerical stability in complex multimodal landscapes. First, a hierarchical guidance mechanism divides the population into multiple groups and constructs a composite direction vector from individual, regional, and global leaders. This guidance is further refined by a dual-layer control scheme: a scheduled mixing strategy that gradually shifts the weights from personal to global experience, and a distance-gated control that attenuates global attraction for groups already close to the current best. Second, population rejuvenation is achieved via Periodic Guided Restart and Opposition-Based Learning, which regenerate poorly performing particles around the global leader and simultaneously probe their opposite locations to restore diversity. Third, a reflective-damped boundary handler preserves velocity continuity while preventing boundary stagnation. Finally, for low-dimensional problems, heterogeneous coefficients, tightened velocity limits, and centroid-based gravitational stabilization jointly prevent trajectory synchronization and oscillatory behavior, yielding smoother convergence and more reliable global optimization performance across diverse benchmark functions and path-planning scenarios. The flow of the algorithm is shown in Fig.~\ref{fig:proposed_algorithm}.

\section{Path planning}\label{sec4}
\subsection{Algorithm performance test}\label{subsec2}

\begin{figure}[htbp]
    \centering
    \includegraphics[width=\textwidth]{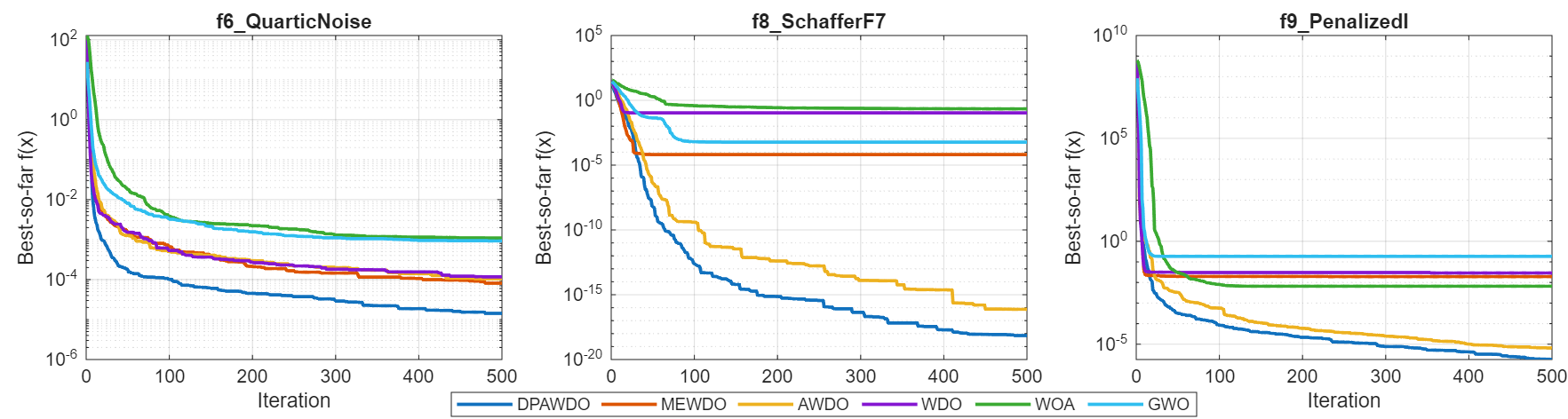}
    \includegraphics[width=\textwidth]{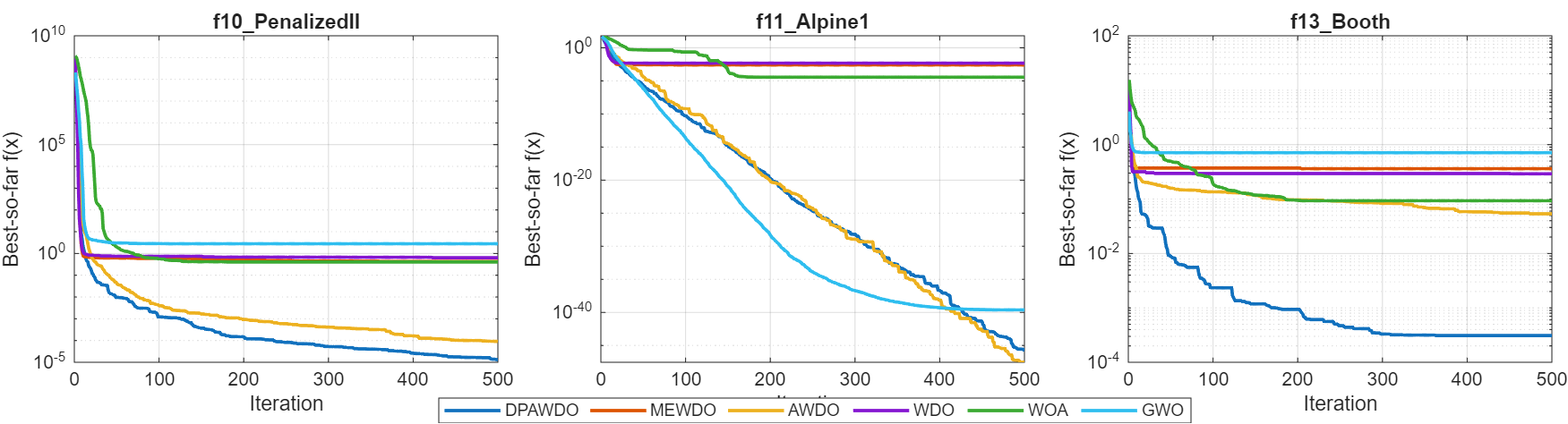}
    \includegraphics[width=\textwidth]{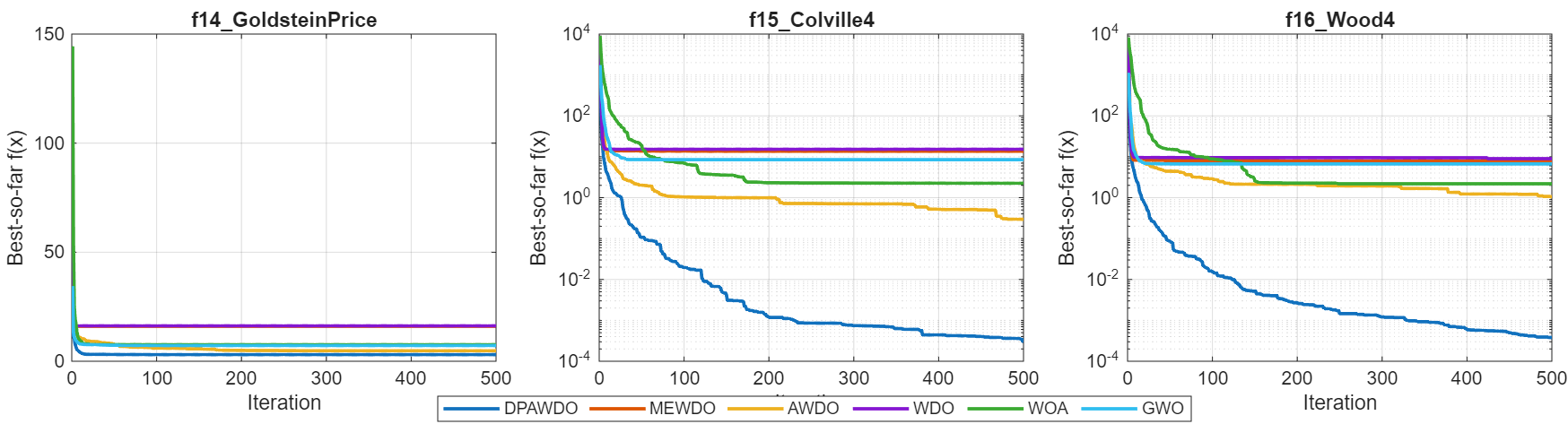}
    \caption{Convergence curves of the comparative experiments on benchmark functions.}
    \label{fig:algorithm_comparison}
\end{figure}

To evaluate the effectiveness of the proposed improved algorithm, a comparative study was conducted against five well-established metaheuristic algorithms: GWO\cite{mirjalili2014gwo}, WOA \cite{mirjalili2016woa}, WDO \cite{bayraktar2010wdo}, AWDO \cite{das2013awdo}, and MEWDO \cite{mewdo2025robotpath}. GWO and WOA represent popular swarm-based optimizers frequently employed as performance benchmarks, while WDO and its adaptive variants are based on atmospheric dynamics. MEWDO further integrates multi-strategy enhancements to improve convergence reliability. All algorithms were configured with a population size of 50 and a maximum of 500 iterations, consistent with common benchmarking practices in evolutionary computation \cite{engelbrecht2007, yang2021book}. To minimize stochastic effects and ensure statistical reliability, each algorithm was independently executed 30 times. The mean, standard deviation, and best (optimal) fitness values were recorded to comprehensively assess their optimization performance. The experimental results are presented in Table~\ref{tab:benchmarkfunctions}, while the average convergence curves of all algorithms on the benchmark functions are illustrated in Fig.~\ref{fig:algorithm_comparison}.

The Wilcoxon rank-sum test at a 5\% significance level was employed to provide a more intuitive comparison of the algorithms’ performance, a standard procedure in metaheuristic statistical validation \cite{yang2021book}. In this analysis, a “+” indicates that the proposed MAWDO algorithm performs significantly better than the compared algorithm, a “–” denotes inferior performance, and a “=” represents no significant difference. All experiments were conducted under a Windows 10 environment with an Intel Core i7-10750H CPU @ 2.6 GHz, using MATLAB 2025a for simulation.

As shown in Table~\ref{tab:benchmarkfunctions}, MAWDO achieved superior fitness values on eight benchmark functions (F6, F8, F10, F11, F13, F14, F15, and F16) compared with the other five algorithms. For F1-F4, WDO and its variants exhibited similar performance, indicating that these unimodal functions are already well-optimized by the baseline methods. Although MAWDO yielded a slightly inferior best result on F9 compared with WDO, it achieved better mean and standard deviation values, reflecting improved stability. For F7-F11, the results demonstrate that MAWDO effectively enhances AWDO’s capability in handling multimodal functions, while for F12-F16, it strengthens performance on low-dimensional problems, consistent with prior findings on enhanced WDO variants.

\begin{sidewaystable*}[htbp]
\centering
\caption{Benchmark Test Functions (F1-F16)}
\label{tab:benchmarkfunctions}
\renewcommand{\arraystretch}{1.35}
\setlength{\tabcolsep}{5pt}

\resizebox{\textheight}{!}{%
\begin{tabular}{c p{12cm} c c c c}
\hline
\textbf{NO.} & \textbf{Function} & \textbf{Dim.} &
\textbf{Search space} & \textbf{minf} & \textbf{Type} \\
\hline

F1  & $f(x)=\sum_{i=1}^{d} x_i^2$ & 30 & $[-100,100]$ & 0  & Unimodal \\
F2  & $f(x)=\sum_{i=1}^{n} |x_i| + \prod_{i=1}^{n} |x_i|$ & 30 & $[-100,100]$ & 0  & Unimodal \\
F3  & $f(x)=\sum_{i=1}^{n}\left( \sum_{j=1}^{i} x_j \right)^2$ & 30 & $[-100,100]$ & 0  & Unimodal \\
F4  & $f(x)=\max_{1\le i\le n} |x_i|$ & 30 & $[-100,100]$ & 0  & Unimodal \\
F5  & $f(x)=\sum_{i=1}^{n-1}\left[ 100(x_{i+1}-x_i^2)^2 + (x_i-1)^2 \right]$
    & 30 & $[-30,30]$ & 0  & Unimodal \\
F6  & $f(x)=\sum_{i=1}^{n}(i\,x_i^4 + \text{random}(0,1))$ & 30 & $[-1.28,1.28]$ & 0 & Unimodal \\
F7  & $f(x)=\sum_{i=1}^{n}\big[ x_i^2 - 10\cos(2\pi x_i) + 10 \big]$
    & 30 & $[-100,100]$ & 0  & Multimodal \\
F8  & $f(x)=\sum_{i=1}^{n} ( s_i + 0.2 s_i^2 ),\ s_i = x_i - x_{i-1}$ 
    & 30 & $[-100,100]$ & 0  & Multimodal \\
F9 & 
$\begin{aligned}
f(x)= & \sum_{i=1}^{n-1} \Big[10\sin^2(\pi y_i)
+ (y_i - 1)^2 (1 + 10\sin^2(\pi y_{i+1}))\Big] \\
& + (y_n - 1)^2 + \sum_{i=1}^n u(x_i)
\end{aligned}$ 
& 30 & $[-50,50]$ & 0 & Multimodal \\

F10 & 
$\begin{aligned}
f(x)= & \frac{\pi}{n}\Big[10\sin^2(\pi y_1)
+ \sum_{i=1}^{n-1} (y_i - 1)^2(1 + 10\sin^2(\pi y_{i+1})) \\
& + (y_n - 1)^2 \Big] + \sum_{i=1}^{n}u(x_i)
\end{aligned}$ 
& 30 & $[-50,50]$ & 0 & Multimodal \\

F11 & $\sum_{i=1}^n [\sin(x_i)+ 0.1x_i^2]$ & 30 & $[-10,10]$ & 0 & Multimodal \\
F12 & $4x_1^2 - 2.1x_1^4 + \frac{x_1^6}{3} + x_1 x_2 - 4x_2^2 + 4x_2^4$
    & 2  & $[-5,5]$ & $-1.03$ & Multimodal \\
F13 & $ -\sum_{i=1}^{4} c_i\, \exp\!\left(-\sum_{j=1}^{6} a_{ij}(x_j - p_{ij})^2\right)$
    & 6  & $[0,1]$ & $-3.32$ & Multimodal \\
F14 & $ -\sum_{i=1}^{5} \frac{1}{(x - a_i)^T (x - a_i) + c_i}$
    & 4  & $[0,10]$ & $-10.15$ & Multimodal \\
F15 & $ -\sum_{i=1}^{7} \frac{1}{(x - a_i)^T (x - a_i) + c_i}$
    & 4  & $[0,10]$ & $-10.40$ & Multimodal \\
F16 & $ -\sum_{i=1}^{10} \frac{1}{(x - a_i)^T (x - a_i) + c_i}$
    & 4  & $[0,10]$ & $-10.54$ & Multimodal \\
\hline
\end{tabular}%
} 

\end{sidewaystable*}

In summary, MAWDO consistently achieved competitive or superior results across the 16 benchmark functions. Even in cases where it did not attain the absolute best fitness value, it maintained favorable convergence behavior and robustness, underscoring its strong overall optimization capability.

\begin{center} 
\begin{longtable}{cccccccc}
\caption{Result of each algorithm of these benchmark functions}\label{tab:results}\\
\toprule
Function & Index & GWO & WOA & WDO & AWDO & MEWDO & MAWDO \\
\midrule
\endfirsthead

\multicolumn{8}{c}%
{{\bfseries \tablename\ \thetable{} -- continued from previous page}} \\
\toprule
Function & Index & GWO & WOA & WDO & AWDO & MEWDO & MAWDO \\
\midrule
\endhead

\midrule \multicolumn{8}{r}{{Continued on next page}} \\
\endfoot

\bottomrule
\endlastfoot

\multirow{4}{*}{F1} & Best & 1.23E-72& 1.77E-85 & 0 & 0 & 0 & 0 \\
& Ave & 2.78E-63 & 8.79E-74 & 0 & 0 & 0 & 0 \\
& Std & 1.01E-62 & 4.67E-73 & 0 & 0 & 0 & 0 \\
& P & + & + & = & = & = & \\

\multirow{4}{*}{F2} & Best & 8.35E-43 & 5.01E-55 & 0 & 0 & 0 & 0 \\
& Ave & 7.52E-39 & 6.36E-48 & 0 & 0 & 0 & 0 \\
& Std & 3.15E-38 & 3.20E-47 & 0 & 0 & 0 & 0 \\
& P & + & + & = & = & = & \\

\multirow{4}{*}{F3} & Best & 1.91E-16 & 1.80E-06 & 0 & 0 & 0 & 0 \\
& Ave & 3.34E-05 & 6275.444 & 0 & 0 & 0 & 0 \\
& Std & 9.78E-05 & 12012.167 & 0 & 0 & 0 & 0 \\
& P & + & + & = & =& = & \\

\multirow{4}{*}{F4} & Best & 2.18E-24 & 1.87E-17 & 0 & 0 & 0 & 0 \\
& Ave & 5.15E-18 & 4.93E-08 & 0 & 0 & 0 & 0 \\
& Std & 2.39E-17 & 2.07E-07 & 0 & 0 & 0 & 0 \\
& P & + & + & = & =& = & \\

\multirow{4}{*}{F5} & Best & 2.87E+01 & 7.55E-02 & 2.87E+01 & 4.12E-04 & 2.87E+01 & 1.37E-02 \\
& Ave & 2.88E+01 & 1.91E+01 & 2.87E+01 & 2.49E+01 & 2.87E+01 & 1.93E+01 \\
& Std & 3.53E-02 & 1.27E+01 & 8.49E-03 & 9.75E+00 & 6.64E-03 & 1.32E+01 \\
& P & + & + & + & - & + &  \\

\multirow{4}{*}{F6} & Best & 2.06E-05 & 5.35E-06 & 7.54E-06 & 5.73E-06 & 4.46E-06 & 6.06E-07 \\
& Ave & 9.29E-04 & 1.09E-03 & 1.17E-04 & 1.05E-04 & 8.15E-05 & 1.43E-05 \\
& Std & 8.60E-04 & 1.51E-03 & 1.18E-04 & 9.16E-05 & 7.97E-05 & 1.22E-05 \\
& P & + & + & + & + & = &  \\

\multirow{4}{*}{F7} & Best & 0 & 0 & 1.60E+00 & 0 & 0 & 0 \\
& Ave & 0 & 0 & 4.16E+00 & 0 & 3.94E+00 & 0 \\
& Std & 0 & 0 & 7.69E-01 & 0 & 8.63E-01 & 0 \\
& P & = & = & + & = & + &  \\

\multirow{4}{*}{F8} & Best & 3.76E-43 & 7.72E-54 & 3.64E-17 & 2.18E-21 & 8.44E-58 & 4.94E-25 \\
& Ave & 5.89E-04 & 2.17E-01 & 1.07E-01 & 7.56E-17 & 6.52E-05 & 7.05E-19 \\
& Std & 3.15E-03 & 7.48E-01 & 4.33E-01 & 1.80E-16 & 1.63E-04 & 1.54E-18 \\
& P & + & + & + & =& + &  \\

\multirow{4}{*}{F9} & Best & 3.48E-02 & 3.40E-04 & 5.96E-04 & 5.10E-08 & 2.19E-03 & 2.38E-07 \\
& Ave & 1.85E-01 & 6.57E-03 & 2.91E-02 & 6.42E-06 & 1.91E-02 & 1.80E-06 \\
& Std & 1.18E-01 & 6.14E-03 & 4.86E-02 & 7.86E-06 & 2.12E-02 & 1.42E-06 \\
& P & + & + & + & - & + &  \\

\multirow{4}{*}{F10} & Best & 1.81E+00 & 6.85E-04 & 1.03E-02 & 8.23E-06 & 6.63E-03 & 8.60E-08 \\
& Ave & 2.80E+00 & 4.07E-01 & 6.38E-01 & 9.05E-05 & 4.30E-01 & 1.33E-05 \\
& Std & 3.53E-01 & 6.33E-01 & 6.68E-01 & 1.15E-04 & 4.54E-01 & 1.48E-05 \\
& P & + & + & + & + & + &  \\

\multirow{4}{*}{F11} & Best & 3.33E-45 & 3.21E-59 & 1.28E-149 & 3.32E-59 & 9.69E-150 & 7.41E-59 \\
& Ave & 2.38E-40 & 3.65E-05 & 4.73E-03 & 2.85E-48 & 2.58E-03 & 1.39E-46 \\
& Std & 1.07E-39 & 1.97E-04 & 1.10E-02 & 1.53E-47 & 9.72E-03 & 6.80E-46 \\
& P & + & + & + & $\approx$ & + &  \\

\multirow{4}{*}{F12} & Best & -1.00E+00 & -1.00E+00 & -9.96E-01 & -1.00E+00 & -1.00E+00 & -1.00E+00 \\
& Ave & -7.38E-01 & -8.80E-01 & -8.42E-01 & -1.00E+00 & -8.74E-01 & -9.99E-01 \\
& Std & 4.17E-01 & 2.98E-01 & 1.67E-01 & 1.42E-04 & 1.45E-01 & 9.97E-04 \\
& P & + & + & + & + & + &  \\

\multirow{4}{*}{F13} & Best & 6.09E-07 & 1.26E-05 & 4.33E-03 & 2.93E-05 & 1.15E-02 & 2.28E-05 \\
& Ave & 7.10E-01 & 9.30E-02 & 2.91E-01 & 5.36E-02 & 3.61E-01 & 3.10E-04 \\
& Std & 1.45E+00 & 2.45E-01 & 3.60E-01 & 2.29E-01 & 3.73E-01 & 3.28E-04 \\
& P & = & + & + & = & + &  \\

\multirow{4}{*}{F14} & Best & 3.00E+00 & 3.00E+00 & 3.12E+00 & 3.00E+00 & 3.21E+00 & 3.00E+00 \\
& Ave & 7.21E+00 & 7.61E+00 & 1.62E+01 & 4.81E+00 & 1.60E+01 & 3.01E+00 \\
& Std & 9.24E+00 & 1.03E+01 & 1.01E+01 & 6.74E+00 & 1.06E+01 & 9.37E-03 \\
& P & + & + & + & + & + &  \\

\multirow{4}{*}{F15} & Best & 2.42E-01 & 3.29E-03 & 2.47E-01 & 2.88E-05 & 3.93E-01 & 9.84E-06 \\
& Ave & 8.50E+00 & 2.26E+00 & 1.53E+01 & 2.96E-01 & 1.36E+01 & 2.98E-04 \\
& Std & 7.25E+00 & 4.09E+00 & 8.45E+00 & 1.10E+00 & 9.40E+00 & 2.58E-04 \\
& P & + & + & + & + & + &  \\
% ...
\multirow{4}{*}{F16} & Best & 1.89E-01 & 6.43E-06 & 1.65E-01 & 1.81E-05 & 9.02E-02 & 2.54E-05 \\
& Ave & 6.67E+00 & 2.18E+00 & 8.97E+00 & 1.05E+00 & 7.80E+00 & 3.81E-04 \\
& Std & 8.58E+00 & 5.24E+00 & 8.34E+00 & 2.90E+00 & 6.79E+00 & 4.33E-04 \\
& P & + & + & + & + & + &  \\

\end{longtable}
\end{center}

\subsection{Ablation experiment}\label{subsec2}

Ablation experiments were conducted to evaluate the contribution of each enhancement component within the proposed algorithm. Specifically, AWDO equipped with multi-group parallelism is denoted as MAWDO-1, AWDO incorporating scheduled triple-guidance as MAWDO-2, AWDO utilizing periodic guided restart as MAWDO-3, and AWDO integrated with low-dimensional stabilization as MAWDO-4. All four variants were tested on the same set of 16 benchmark functions under identical experimental settings, including population size, number of iterations, number of runs, and random seeds. The comparative results are summarized in Table~\ref{tab:ablation}.

\begin{table}[htbp]
\centering
\caption{Result of ablation experiment}
\label{tab:ablation}
\begin{tabular}{ccccccc}
\toprule
Function & Index & MAWDO-1& MAWDO-2& MAWDO-3& MAWDO-4& MAWDO \\
\midrule
\multirow{4}{*}{F1} & Best & 0 & 0 & 0 & 0 & 0 \\
& Ave & 0 & 0 & 0 & 0 & 0 \\
& Std & 0 & 0 & 0 & 0 & 0 \\

\multirow{4}{*}{F2} & Best & 0 & 0 & 0 & 0 & 0 \\
& Ave & 0 & 0 & 0 & 0 & 0 \\
& Std & 0 & 0 & 0 & 0 & 0 \\

\multirow{4}{*}{F3} & Best & 0 & 0 & 0 & 0 & 0 \\
& Ave & 0 & 0 & 0 & 0 & 0 \\
& Std & 0 & 0 & 0 & 0 & 0 \\

\multirow{4}{*}{F4} & Best & 0 & 0 & 0 & 0 & 0 \\
& Ave & 0 & 0 & 0 & 0 & 0 \\
& Std & 0 & 0 & 0 & 0 & 0 \\

\multirow{4}{*}{F5} & Best & 4.68E-02 & 2.87E+01 & 9.97E-01 & 4.69E-01 & 1.37E-02 \\
& Ave & 1.91E+01 & 2.87E+01 & 2.64E+01 & 2.64E+01 & 1.93E+01 \\
& Std & 1.27E+01 & 1.82E-03 & 7.14E+00 & 7.28E+00 & 1.32E+01 \\

\multirow{4}{*}{F6} & Best & 1.23E-06 & 5.16E-06 & 1.83E-05 & 5.16E-06 & 6.06E-07 \\
& Ave & 1.78E-05 & 1.02E-04 & 1.14E-04 & 1.11E-04 & 1.43E-05 \\
& Std & 1.72E-05 & 9.18E-05 & 9.42E-05 & 9.54E-05 & 1.22E-05 \\

\multirow{4}{*}{F7} & Best & 0 & 0 & 0 & 0 & 0 \\
& Ave & 0 & 0 & 0 & 0 & 0 \\
& Std & 0 & 0 & 0 & 0 & 0 \\

\multirow{4}{*}{F8} & Best & 1.30E-23 & 6.41E-24 & 4.16E-20 & 9.41E-21 & 4.94E-25 \\
& Ave & 3.39E-19 & 6.28E-17 & 3.81E-16 & 2.55E-16 & 7.05E-19 \\
& Std & 8.12E-19 & 1.39E-16 & 8.43E-16 & 6.18E-16 & 1.54E-18 \\

\multirow{4}{*}{F9} & Best & 4.71E-08 & 5.68E-07 & 5.08E-07 & 7.47E-08 & 2.38E-07 \\
& Ave & 2.23E-06 & 1.67E-05 & 3.34E-05 & 2.94E-05 & 1.80E-06 \\
& Std & 2.62E-06 & 1.89E-05 & 5.76E-05 & 3.44E-05 & 1.42E-06 \\

\multirow{4}{*}{F10} & Best & 1.08E-06 & 1.16E-05 & 5.62E-06 & 1.56E-05 & 8.60E-08 \\
& Ave & 4.81E-05 & 1.00E-01 & 1.60E-01 & 2.01E-01 & 1.33E-05 \\
& Std & 5.33E-05 & 5.38E-01 & 6.15E-01 & 7.47E-01 & 1.48E-05 \\

\multirow{4}{*}{F11} & Best & 1.05E-62 & 4.57E-60 & 1.24E-57 & 4.60E-63 & 7.41E-59 \\
& Ave & 1.06E-49 & 5.57E-48 & 7.98E-47 & 2.53E-50 & 1.39E-46 \\
& Std & 5.56E-49 & 2.43E-47 & 3.25E-46 & 9.42E-50 & 6.80E-46 \\

\multirow{4}{*}{F12} & Best & -1.00E+00 & -1.00E+00 & -1.00E+00 & -9.98E-01 & -1.00E+00 \\
& Ave & -1.00E+00 & -1.00E+00 & -1.00E+00 & -2.02E-01 & -9.99E-01 \\
& Std & 5.88E-05 & 1.18E-04 & 8.21E-04 & 3.65E-01 & 9.97E-04 \\

\multirow{4}{*}{F13} & Best & 8.49E-06 & 1.02E-03 & 1.08E-04 & 1.15E-12 & 2.28E-05 \\
& Ave & 1.68E-03 & 8.87E-02 & 1.23E-02 & 8.41E-02 & 3.10E-04 \\
& Std & 2.37E-03 & 1.59E-01 & 1.74E-02 & 3.47E-01 & 3.28E-04 \\

\multirow{4}{*}{F14} & Best & 3.00E+00 & 3.00E+00 & 3.00E+00 & 3.02E+00 & 3.00E+00 \\
& Ave & 3.03E+00 & 6.35E+00 & 3.12E+00 & 9.41E+00 & 3.01E+00 \\
& Std & 2.47E-02 & 7.92E+00 & 2.08E-01 & 7.54E+00 & 9.37E-03 \\

\multirow{4}{*}{F15} & Best & 9.68E-06 & 4.38E-05 & 1.20E-04 & 4.32E-04 & 9.84E-06 \\
& Ave & 6.55E-04 & 1.36E-01 & 1.47E-02 & 1.39E-02 & 2.98E-04 \\
& Std & 9.51E-04 & 7.05E-01 & 2.24E-02 & 1.23E-02 & 2.58E-04 \\

\multirow{4}{*}{F16} & Best & 6.76E-06 & 6.86E-05 & 5.38E-05 & 1.03E-04 & 2.54E-05 \\
& Ave & 4.00E-04 & 7.99E-03 & 1.56E-02 & 8.67E-03 & 3.81E-04 \\
& Std & 5.12E-04 & 1.29E-02 & 1.59E-02 & 1.18E-02 &  \\
\bottomrule
\end{tabular}
\end{table}

\subsection{Cost function}\label{subsec2}
In dynamic robot path planning, the cost function is formulated to evaluate the quality of the generated path and to guide the optimizer toward an optimal trajectory. It integrates several key objectives, namely, path length, obstacle avoidance, and path smoothness, which collectively determine the feasibility, safety, and efficiency of the robot’s motion.
The cost function in this paper is formulated as follows:
\begin{equation}
    C = \alpha C_1 + \beta C_2 + \lambda C_3
\end{equation}
where C represents the total cost of the path, and the three sub-costs $C_1$, $C_2$, $C_3$ denote the path length cost, obstacle avoidance cost, and turning smoothness cost, respectively. The weighting coefficients $\alpha$, $\beta$, and $\lambda$ are normalized factors such that:
\begin{equation}
    \alpha+\beta+\lambda=1
\end{equation}
where C,$C_1$,$C_2$, and $C_3$ represent total cost, path distance cost, obstacle avoidance cost, and turning cost respectively. $\alpha$,  $\beta$ , and $\lambda$  are proportion coefficients where their sum equals 1. 

\subsubsection{Path distance cost C1}\label{subsubsec2}
In robot path planning, the path length cost represents a fundamental criterion for evaluating the overall performance of a generated trajectory. Minimizing the total path distance enhances the robot’s navigational efficiency while reducing both travel time and energy consumption. This cost component quantifies the total Euclidean distance of the planned route, computed as the cumulative sum of the segmental distances between consecutive waypoints along the robot's trajectory.
\begin{equation}
    C_1 = \sum_{i=1}^{N_p - 1} \sqrt{(x_{i+1} - x_i)^2 + (y_{i+1} - y_i)^2}
\end{equation}
Where $N_p$ denotes the number of path waypoints, and $(x_i, y_i)$ represents the coordinates of the $i^{th}$ waypoint.

\subsubsection{Obstacle avoidance cost C2}\label{subsubsec2}
The coefficient $c_2$ serves not only to prevent collisions but also to guide the optimization process toward safer regions of the search space. By assigning a sufficiently large penalty value, the optimizer learns that even minor collisions result in a substantial increase in the objective cost, thereby discouraging candidate solutions from approaching obstacle regions. In this study, obstacle intersections are identified through a segment intersection test, in which the number of collisions between the planned path segments and obstacle boundaries is computed. Paths exhibiting any intersection with obstacles are penalized heavily to ensure robust obstacle avoidance and promote the generation of feasible trajectories:
\begin{equation}
    C_2 = k \times n_{collision}^4
\end{equation}
Where k is a scaling factor, and $n_{collision}$ is the number of obstacle intersections per path.

\subsubsection{Turning smoothness cost C3}\label{subsubsec2}
Sharp turns necessitate abrupt acceleration and deceleration, which consequently increase motor torque and power consumption, accelerate battery depletion, and induce mechanical wear on rotors or wheels. By minimizing the coefficient $c_3$, the robot maintains a smoother velocity profile and reduced energy consumption, thereby enhancing mission endurance and extending hardware lifespan, factors that are critical for long-duration or repetitive missions.

To ensure that the robot follows a physically smooth trajectory without abrupt directional changes, this cost component evaluates the cumulative turning angle between consecutive path segments. The cost value increases proportionally with the deviation of each turning angle from a straight-line trajectory, effectively penalizing sharp turns and promoting path smoothness.
\begin{equation}
    C_3 = \sum_{i=2}^{N_p - 1} \left| \theta_i - \pi \right|
\end{equation}
where $\theta_i$ is the turning angle at waypoint $i$.

The MAWDO algorithm minimizes the overall cost $C$ to obtain a dynamically feasible and collision-free flight path. The cost function simultaneously ensures shorter path length, collision avoidance, and smooth maneuvering, which are essential for real-time path planning in dynamic environments.

\subsection{Experiment simulation}\label{subsec2}
To validate the practical applicability of the proposed algorithm, a simulation experiment was conducted on a (366 $\times$ 366)-pixel map containing multiple obstacles. Among these, six black obstacles were dynamic, while two yellow obstacles remained static. The paths generated by five algorithms, GWO, WOA, WDO, AWDO, and the proposed MEWDO, were compared. For each algorithm, the number of groups (G) was set to 8, with each group comprising 170 particles. Each particle was represented as a 40-dimensional vector corresponding to 20 waypoints.

\begin{figure*}[htbp]
\centering

\textbf{GWO} \\
\includegraphics[width=0.11\textwidth]{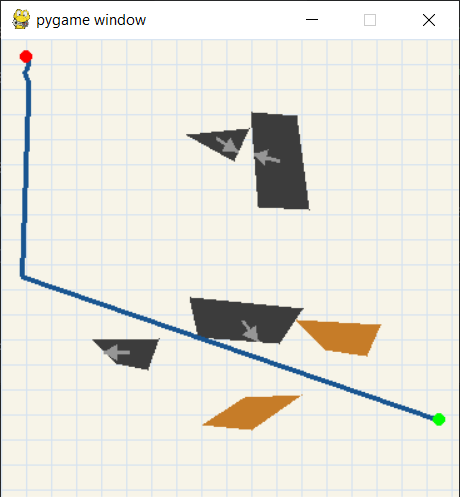}
\includegraphics[width=0.11\textwidth]{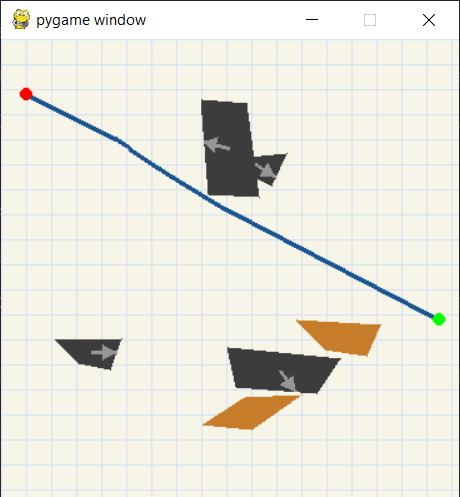}
\includegraphics[width=0.11\textwidth]{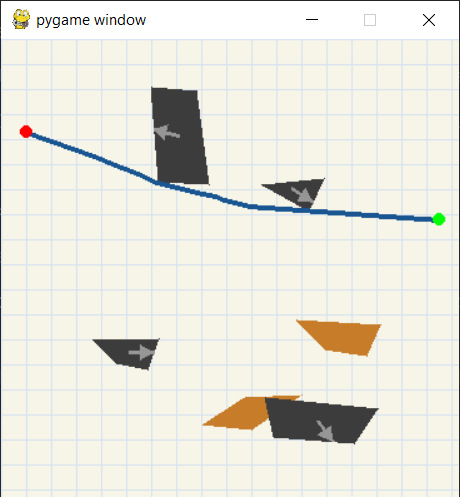}
\includegraphics[width=0.11\textwidth]{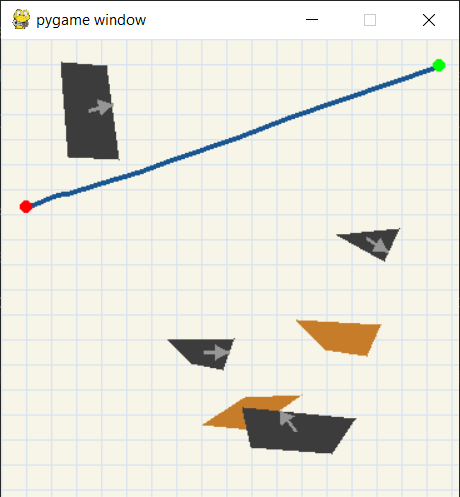}
\includegraphics[width=0.11\textwidth]{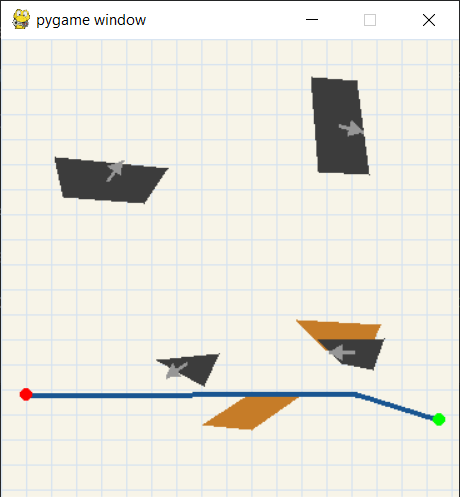}
\includegraphics[width=0.11\textwidth]{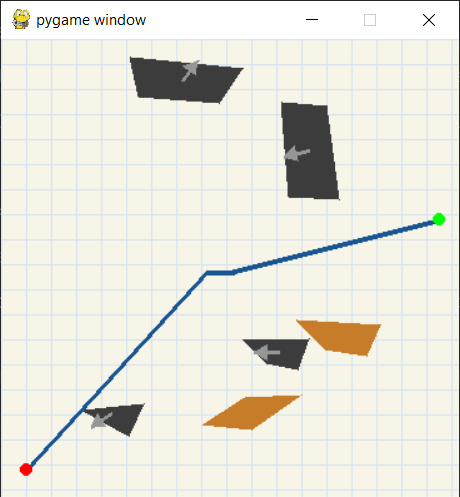}
\includegraphics[width=0.11\textwidth]{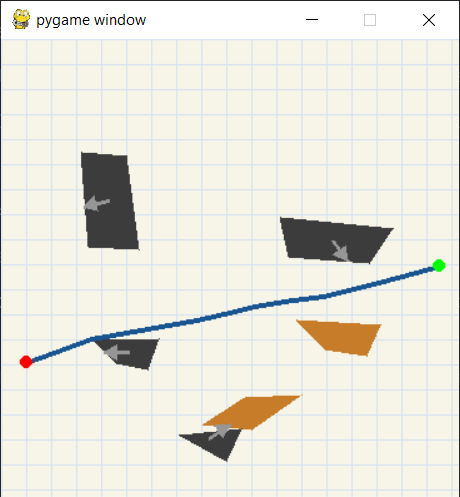}
\includegraphics[width=0.11\textwidth]{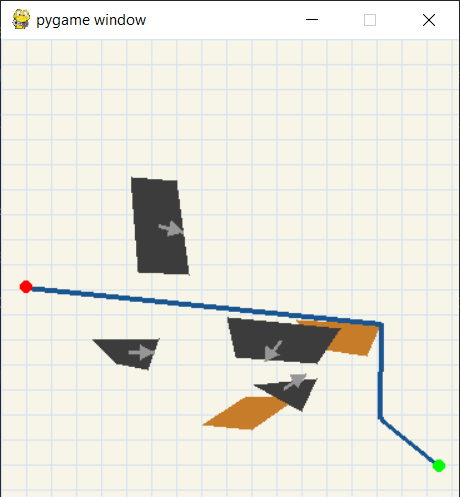}

\vspace{8pt}

\textbf{WOA} \\
\includegraphics[width=0.11\textwidth]{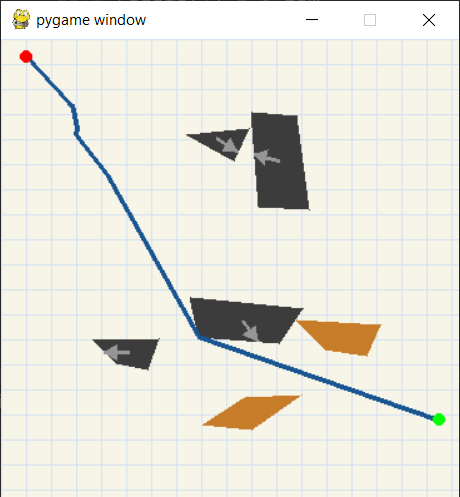}
\includegraphics[width=0.11\textwidth]{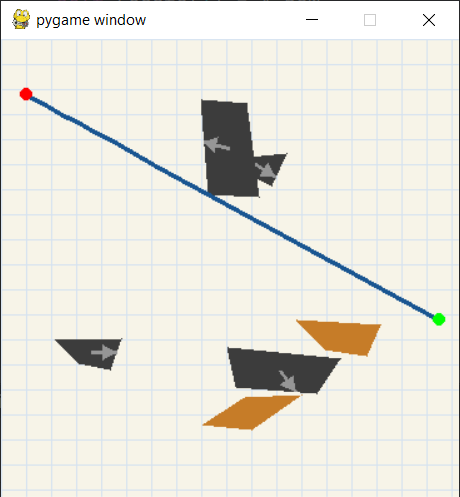}
\includegraphics[width=0.11\textwidth]{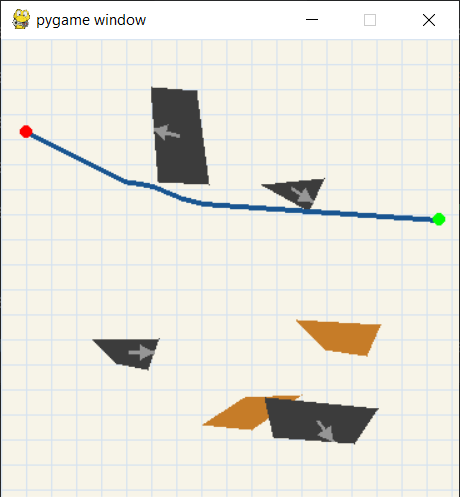}
\includegraphics[width=0.11\textwidth]{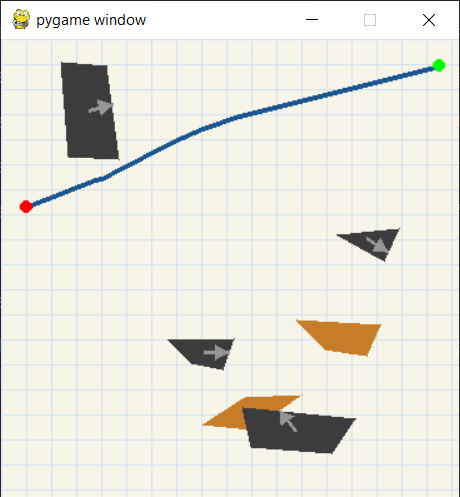}
\includegraphics[width=0.11\textwidth]{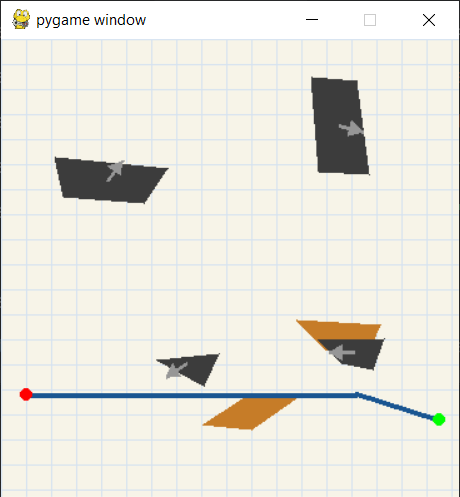}
\includegraphics[width=0.11\textwidth]{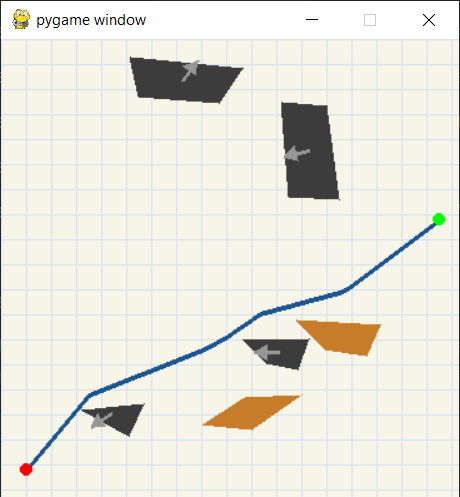}
\includegraphics[width=0.11\textwidth]{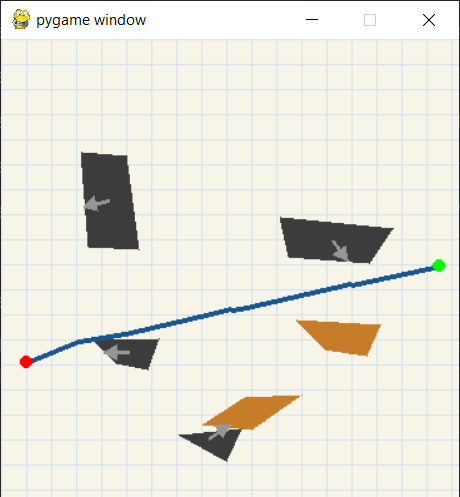}
\includegraphics[width=0.11\textwidth]{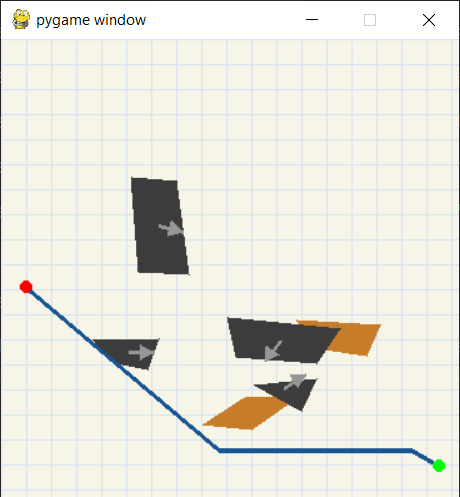}

\vspace{8pt}

\textbf{WDO} \\
\includegraphics[width=0.11\textwidth]{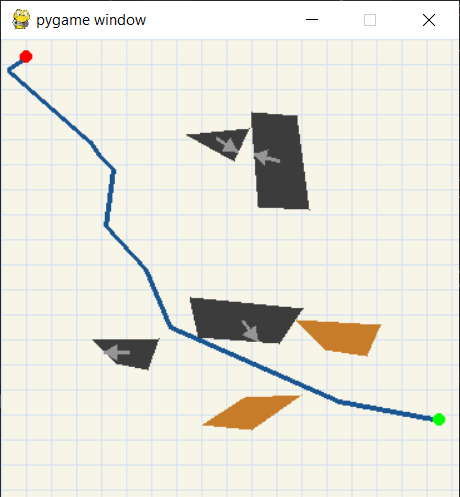}
\includegraphics[width=0.11\textwidth]{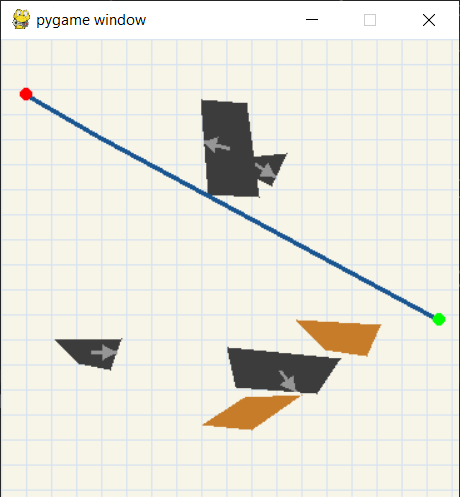}
\includegraphics[width=0.11\textwidth]{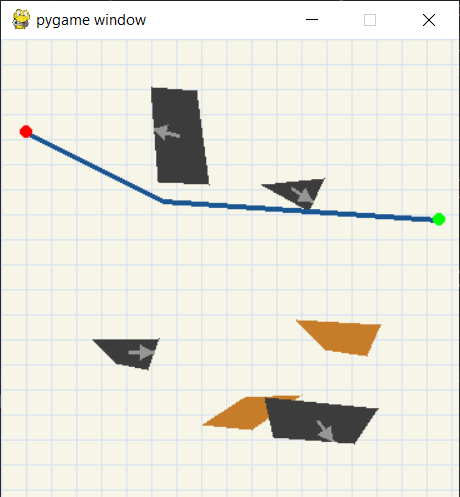}
\includegraphics[width=0.11\textwidth]{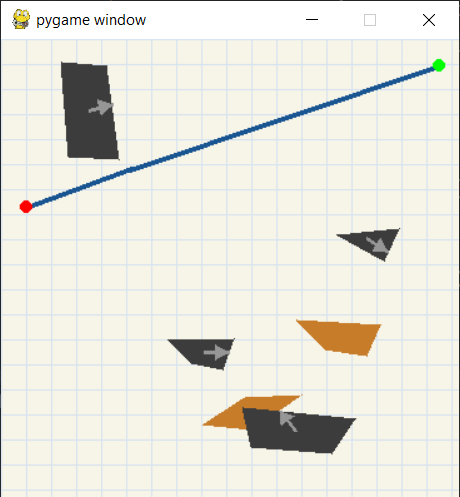}
\includegraphics[width=0.11\textwidth]{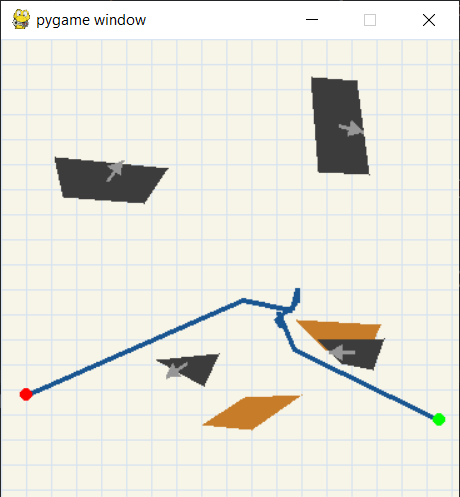}
\includegraphics[width=0.11\textwidth]{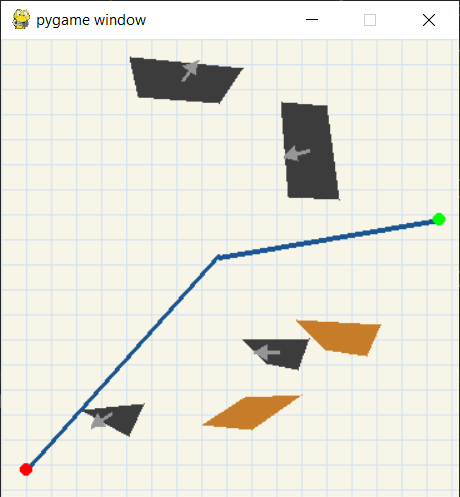}
\includegraphics[width=0.11\textwidth]{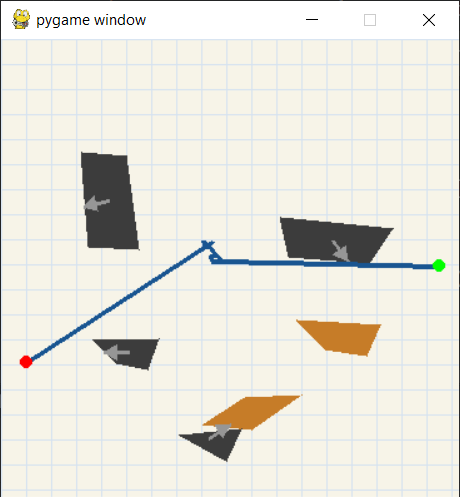}
\includegraphics[width=0.11\textwidth]{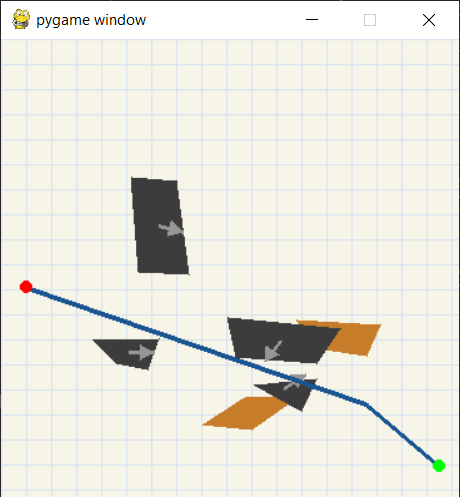}

\vspace{8pt}

\textbf{AWDO} \\
\includegraphics[width=0.11\textwidth]{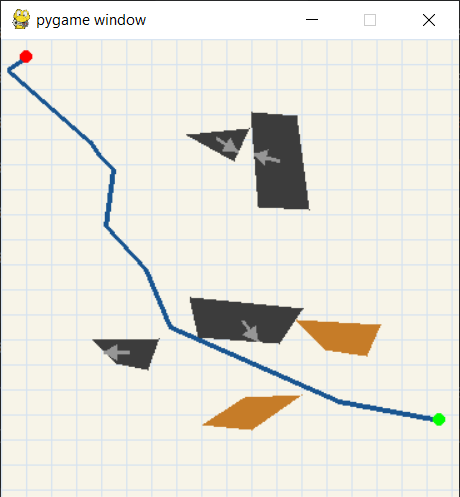}
\includegraphics[width=0.11\textwidth]{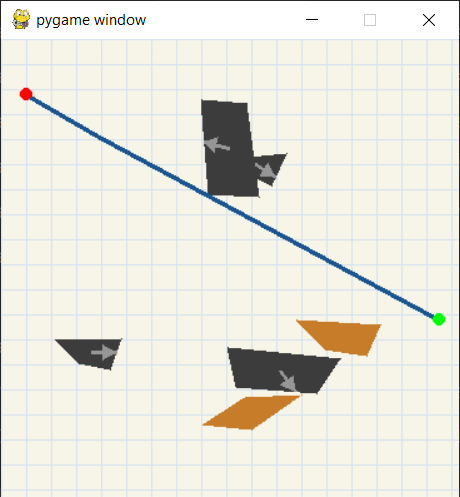}
\includegraphics[width=0.11\textwidth]{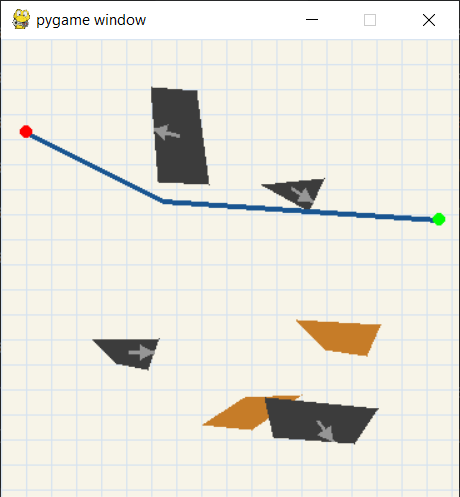}
\includegraphics[width=0.11\textwidth]{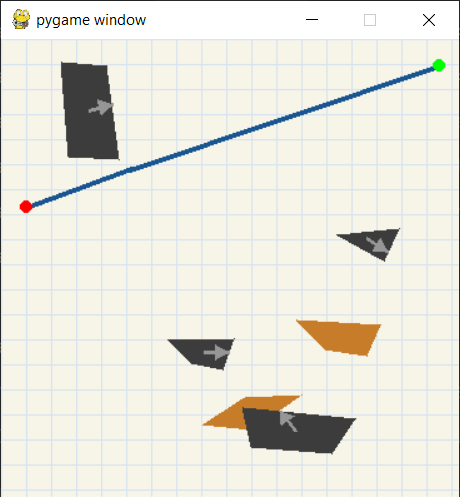}
\includegraphics[width=0.11\textwidth]{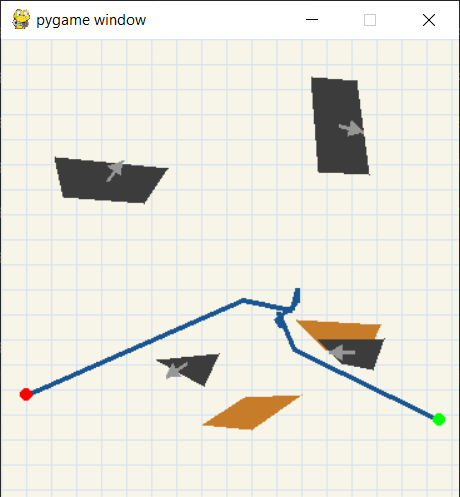}
\includegraphics[width=0.11\textwidth]{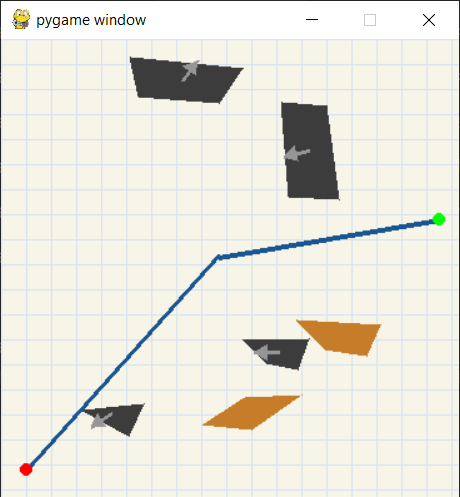}
\includegraphics[width=0.11\textwidth]{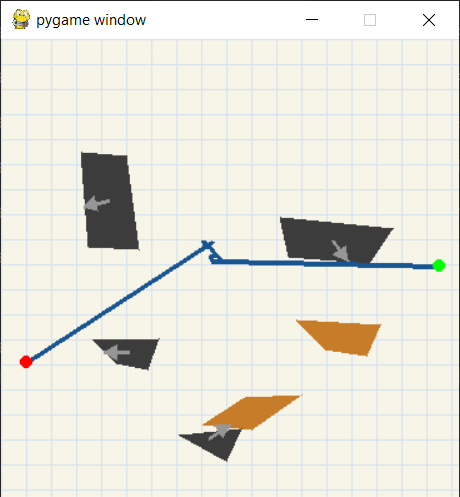}
\includegraphics[width=0.11\textwidth]{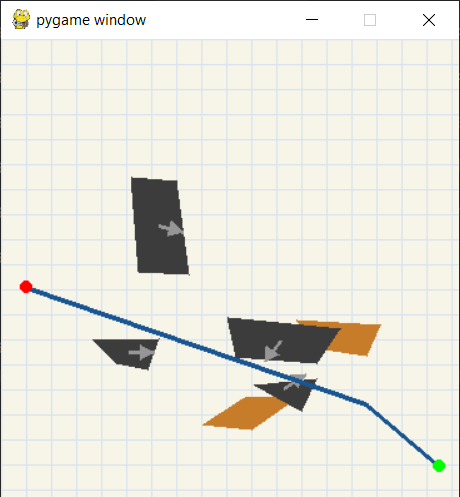}

\vspace{8pt}

\textbf{MEWDO} \\
\includegraphics[width=0.11\textwidth]{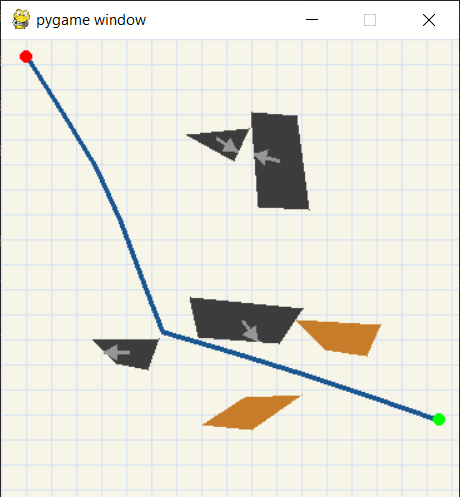}
\includegraphics[width=0.11\textwidth]{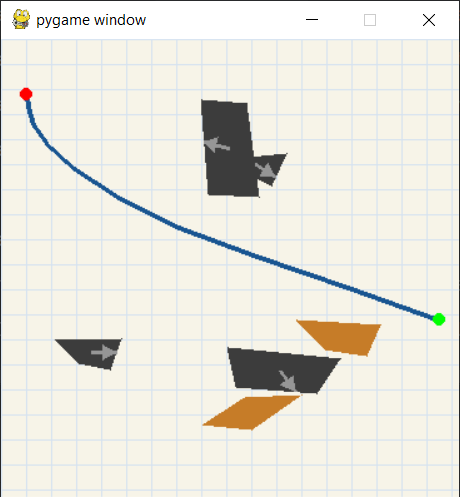}
\includegraphics[width=0.11\textwidth]{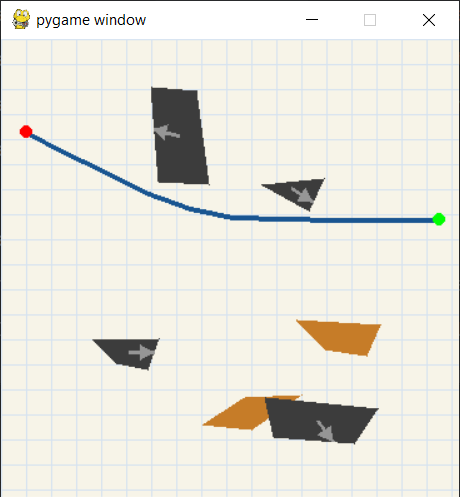}
\includegraphics[width=0.11\textwidth]{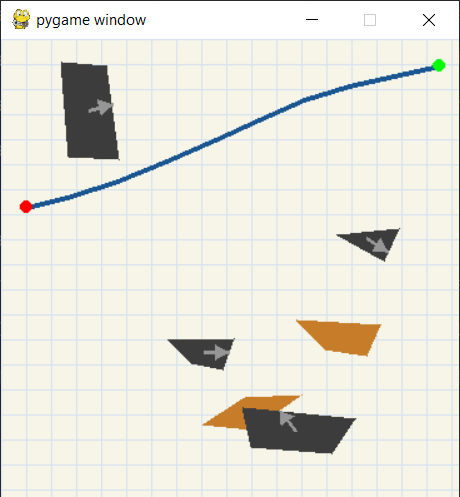}
\includegraphics[width=0.11\textwidth]{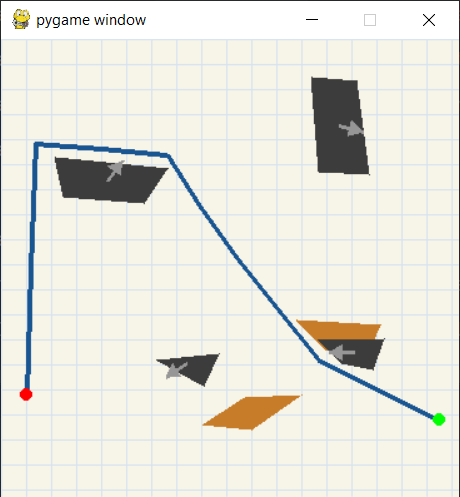}
\includegraphics[width=0.11\textwidth]{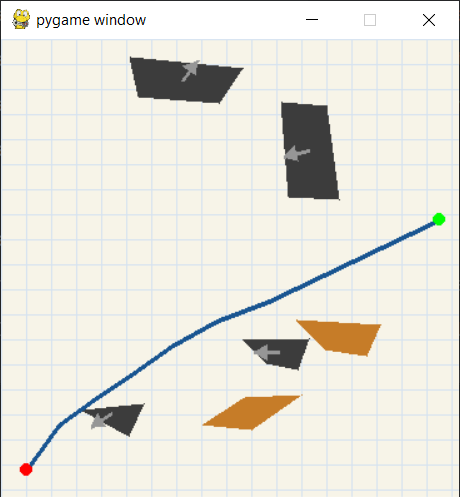}
\includegraphics[width=0.11\textwidth]{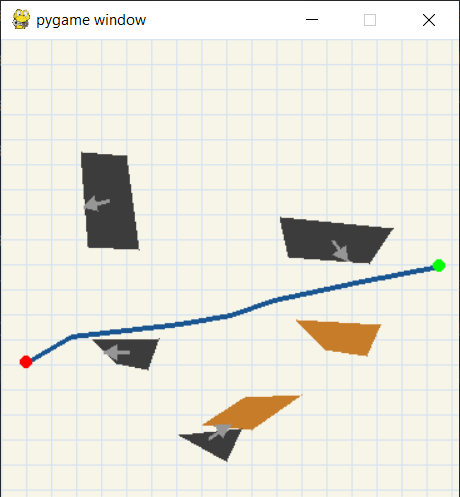}
\includegraphics[width=0.11\textwidth]{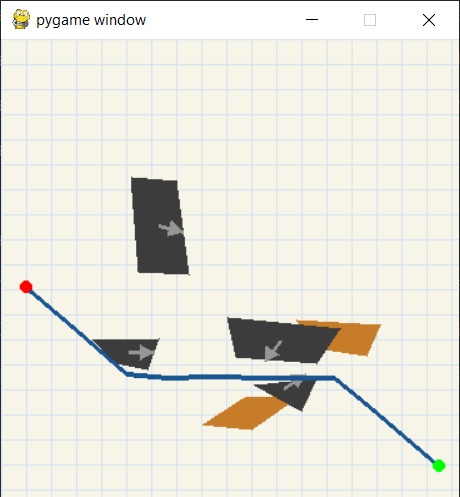}

\vspace{8pt}

\textbf{MAWDO} \\
\includegraphics[width=0.11\textwidth]{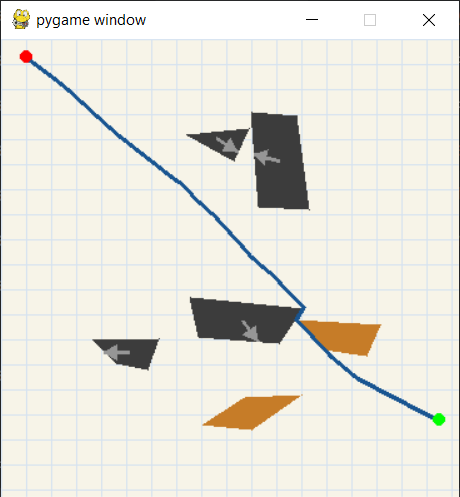}
\includegraphics[width=0.11\textwidth]{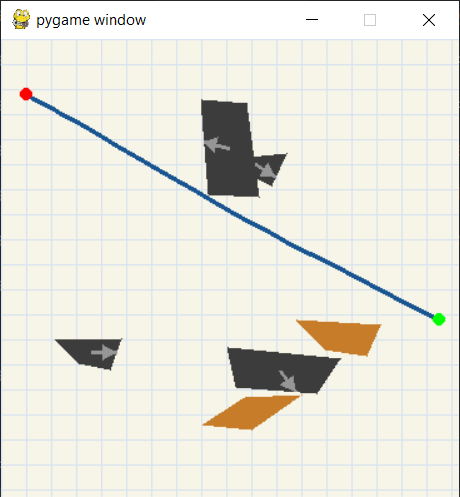}
\includegraphics[width=0.11\textwidth]{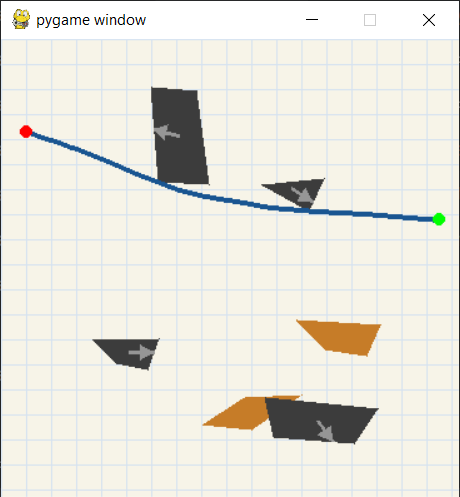}
\includegraphics[width=0.11\textwidth]{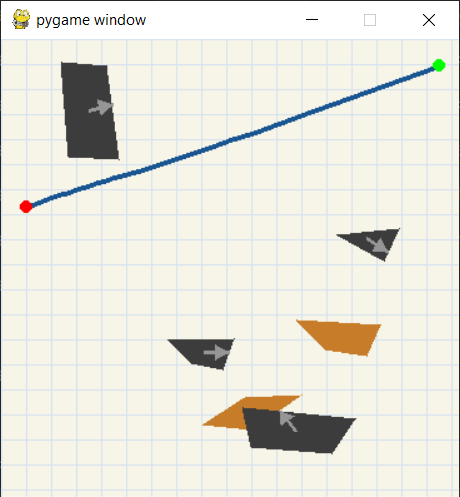}
\includegraphics[width=0.11\textwidth]{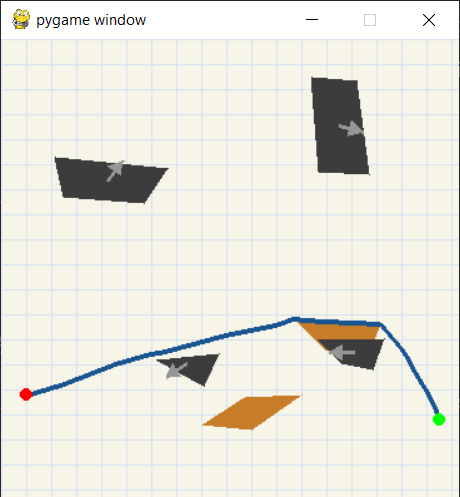}
\includegraphics[width=0.11\textwidth]{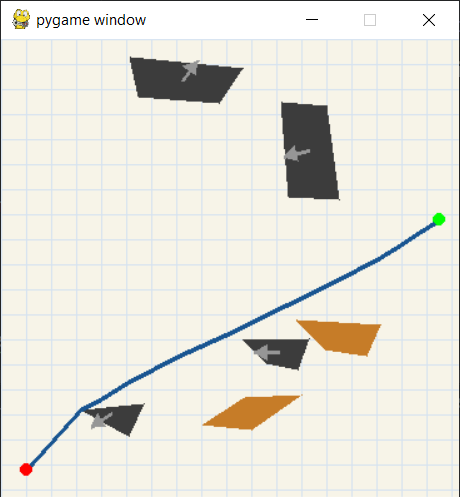}
\includegraphics[width=0.11\textwidth]{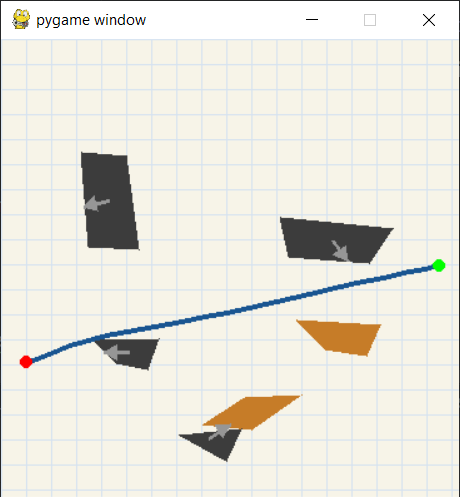}
\includegraphics[width=0.11\textwidth]{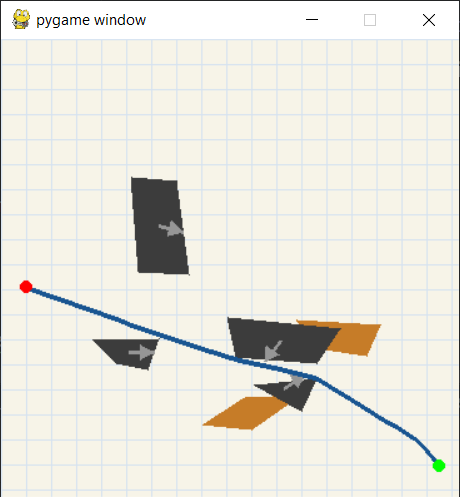}

\caption{Trajectory Comparison of Six Algorithms}
\label{fig:path_maps}

\end{figure*}

\begin{figure}[htbp]
    \centering
    \includegraphics[width=0.5\textwidth]{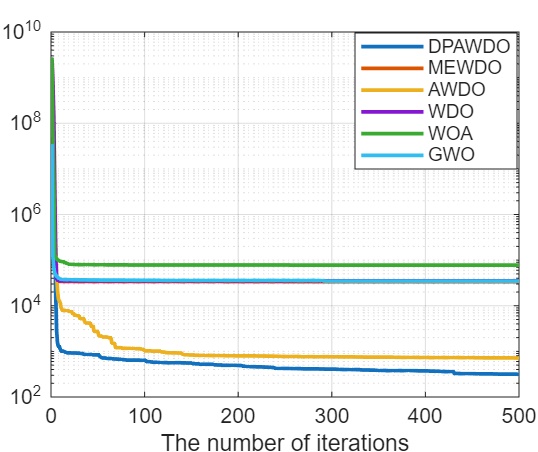}
    \caption{Cost function convergence curve}
    \label{fig:convergence_curve}
\end{figure}

To ensure experimental credibility and minimize stochastic effects, each algorithm was independently executed 50 times. The path length, optimality gap, and smoothness were recorded for statistical analysis. The best path maps and convergence curves are illustrated in Fig.~\ref{fig:path_maps} and Fig.~\ref{fig:convergence_curve}, respectively, while the quantitative performance metrics are summarized in Table~\ref{tab:quantitative_performance}.

As observed in Fig.~\ref{fig:path_maps}, the proposed algorithm produces smoother paths with fewer oscillations, effectively avoiding multiple obstacles. From Fig.~\ref{fig:convergence_curve} and Table~\ref{tab:quantitative_performance}, it can be seen that the proposed method achieves the fastest convergence rate and the smallest optimality gap (1.01). In addition, the paths generated by the proposed algorithm exhibit superior smoothness compared with those produced by AWDO and WDO. From the perspective of path length, significant improvements are also observed over MEWDO and AWDO. In a word, the proposed algorithm demonstrates excellent convergence characteristics and rapid performance in the context of mobile robot path planning.

\begin{table}[htbp]
\centering
\caption{Performance index}
\label{tab:quantitative_performance}
\begin{tabular}{lccc}
\toprule
\textbf{Algorithm} & \textbf{Length (pixels)} & \textbf{Optimality Gap} & \textbf{Smooth} \\
\midrule
DPAWDO  & \textbf{469.28} & \textbf{1.01} & 0.71 \\
MEWDO  & 486.37 & 1.05 & \textbf{0.47} \\
AWDO   & 531.06 & 1.15 & 13.50 \\
WDO    & 551.65 & 1.19 & 15.67 \\
WOA    & 472.72 & 1.02 & 0.81 \\
GWO    & 471.86 & 1.02 & 0.74 \\
\bottomrule
\end{tabular}
\end{table}

\section{Conclusion}\label{sec5}

This work introduced a multi-group, hierarchical-guidance AWDO that integrates a set of low-dimensional stabilization mechanisms, velocity-limit tightening and centroid-based gravitational stabilization, to mitigate premature collapse and improve search reliability in small-D landscapes. Together, these components strengthen exploration–exploitation coordination without sacrificing the algorithm’s simplicity and physical interpretability.  

Extensive benchmarking against GWO, WOA, WDO, AWDO, and MEWDO using 30 independent runs and Wilcoxon rank-sum testing (5\% level) shows that the proposed method attains superior fitness on eight functions (F6, F8, F10, F11, F13, F14, F15, F16) and maintains competitive robustness elsewhere, underscoring both accuracy and stability. Complementary ablation studies further indicate that each architectural element, multi-hierarchical guidance, periodic guided restart, and low-dimensional stabilization, contribute measurably to the final performance profile. 

On a dynamic robot path-planning task featuring a tri-objective cost (path length, obstacle avoidance, smoothness), the proposed algorithm produced smoother, collision-free trajectories and the strongest convergence behavior among all comparators, achieving a path length of 469.28 pixels with an optimality gap of 1.01 and high smoothness, while competing methods yielded longer or less smooth paths. These results confirm practical gains beyond synthetic benchmarks.   

In summary, the proposed multi-hierarchical AWDO framework offers a principled, low-overhead route to improved reliability on multimodal and low-dimensional problems and translates effectively to dynamic planning scenarios. Future work will investigate adaptive scheduling of guidance weights across groups, tighter theory for stability in very high dimensions, and hardware-in-the-loop validations in real-time robotic platforms.

\FloatBarrier
\section*{Declarations}

\textbf{Competing interests} The authors declare no competing interests.\\
\textbf{Conflict of interest} The authors have no relevant financial or non-financial interests to disclose.\\
\textbf{Ethical approval} Not applicable.


\end{document}